\documentclass{article}

\usepackage{natbib}
\PassOptionsToPackage{numbers, compress}{natbib}

\usepackage[preprint]{neurips_2024}

\usepackage[utf8]{inputenc} 
\usepackage[T1]{fontenc}    
\usepackage[hidelinks]{hyperref}       
\usepackage{url}            
\usepackage{booktabs}       
\usepackage{amsfonts}       
\usepackage{nicefrac}       
\usepackage{microtype}      
\usepackage{xcolor}         
\usepackage{amsmath}
\usepackage{amssymb}
\usepackage{amsthm}
\usepackage{hhline}
\usepackage{rotating}
\usepackage{appendix}

\usepackage{amsmath,amsfonts,bm}

\def\eqref#1{equation~\ref{#1}}

\def\1{\bm{1}}

\DeclareMathAlphabet{\mathsfit}{\encodingdefault}{\sfdefault}{m}{sl}
\SetMathAlphabet{\mathsfit}{bold}{\encodingdefault}{\sfdefault}{bx}{n}

\usepackage{url}

\usepackage{inconsolata}

\usepackage{times}
\usepackage{latexsym}
\usepackage{amsfonts}
\usepackage{amsmath}
\usepackage{amssymb}
\usepackage{multicol}
\usepackage{multirow}
\usepackage{xspace}
\usepackage{booktabs}
\usepackage{bbding}
\usepackage{array}
\usepackage{threeparttable}
\usepackage[most]{tcolorbox}
\usepackage{tabularx}
\usepackage{enumitem}
\usepackage[linesnumbered,ruled,vlined]{algorithm2e}
\usepackage{xcolor,colortbl}
\usepackage{setspace}
\usepackage{makecell}
\usepackage{xltabular}
\usepackage{graphicx}
\usepackage{subfig}
\usepackage{wrapfig,lipsum}

\usepackage{cleveref}
\crefformat{section}{\S#2#1#3}
\crefformat{subsection}{\S#2#1#3}
\crefformat{subsubsection}{\S#2#1#3}
\crefrangeformat{section}{\S\S#3#1#4 to~#5#2#6}
\crefmultiformat{section}{\S\S#2#1#3}{ and~#2#1#3}{, #2#1#3}{ and~#2#1#3}
\crefmultiformat{subsection}{\S\S#2#1#3}{ and~#2#1#3}{, #2#1#3}{ and~#2#1#3}
\Crefformat{figure}{#2Fig.~#1#3}
\Crefmultiformat{figure}{Figs.~#2#1#3}{ and~#2#1#3}{, #2#1#3}{ and~#2#1#3}
\Crefformat{table}{#2Tab.~#1#3}
\Crefmultiformat{table}{Tabs.~#2#1#3}{ and~#2#1#3}{, #2#1#3}{ and~#2#1#3}
\Crefformat{appendix}{Appx.~\S#2#1#3}
\crefmultiformat{appendix}{Appx.~\S#2#1#3}{ and~#2#1#3}{, #2#1#3}{ and~#2#1#3}
\crefformat{algorithm}{Alg.~#2#1#3}
\Crefformat{equation}{Eq.~#2#1#3}

\newcommand{\xhdr}[1]{\vspace{0.3em}\noindent{{\bf #1.}}}

\newcommand{\PreserveBackslash}[1]{\let\temp=\\#1\let\\=\temp}
\newcolumntype{C}[1]{>{\PreserveBackslash\centering}p{#1}}

\title{Online Merging Optimizers for Boosting Rewards and Mitigating Tax in Alignment}

\author{
Keming Lu, Bowen Yu, Fei Huang, Yang Fan, Runji Lin, Chang Zhou
\\
Qwen Team, Alibaba Inc. \\
\texttt{\{lukeming.lkm,yubowen.ybw,feihu.hf\}@alibaba-inc.com}\\
\texttt{\{suyang.fy,linrunji.lrj,ericzhou.zc\}@alibaba-inc.com}\\
}

\begin{document}

\maketitle

\begin{abstract}

Effectively aligning Large Language Models (LLMs) with human-centric values while preventing the degradation of abilities acquired through Pre-training and Supervised Fine-tuning (SFT) poses a central challenge in Reinforcement Learning from Human Feedback (RLHF). 
In this paper, we first discover that interpolating RLHF and SFT model parameters can adjust the trade-off between human preference and basic capabilities, thereby reducing the alignment tax at the cost of alignment reward. 
Inspired by this, we propose integrating the RL policy and SFT models at each optimization step in RLHF to continuously regulate the training direction, introducing the Online Merging Optimizer. 
Specifically,  we merge gradients with the parameter differences between SFT and pretrained models, effectively steering the gradient towards maximizing rewards in the direction of SFT optimization.
We demonstrate that our optimizer works well with different LLM families, such as Qwen and LLaMA, across various model sizes ranging from 1.8B to 8B, various RLHF algorithms like DPO and KTO, and existing model merging methods. 
It significantly enhances alignment reward while mitigating alignment tax, achieving higher overall performance across 14 benchmarks.\footnote{Codes can be found at \url{https://github.com/QwenLM/online_merging_optimizers}}

\end{abstract}
\section{Introduction}

Reinforcement Learning From Human Feedback (RLHF) has propelled the success of the most recent wave of generative AI models~\citep{ouyang2022training,Bai2022ConstitutionalAH}.
While RLHF garners the bonus of steering large language models (LLMs) to meet human expectations, past studies have highlighted that this approach can inadvertently lead to forgetting the diverse abilities that the LLMs have already mastered through pre-training and supervised finetuning~(SFT) \citep{bai2022training,zheng2023secrets,lin2024mitigating,dong2024abilities}.
This is evident in their declining performance on certain public language benchmarks and a loss of fundamental language skills.
Additionally, responses tend to exhibit unexpected code-switching and get significantly longer~\citep {park2024disentangling}.
These undesirable side effects are commonly referred to as the ``alignment tax''~\citep{dong2023raft,sun2024principle}.

Ideally,  an optimal RLHF strategy should maintain the bonuses of alignment while avoiding the associated tax, striving to maximize rewards while minimizing forgetting.
Relying on the linear mode connectivity of neural networks~\citep{garipov2018loss,frankle2020linear,entezari2021role}, the trade-off in model capabilities can be succinctly described as the interpolation of model parameters~\citep{garipov2018loss,ilharco2022editing,zheng2024weak}.
Studies have shown that combining different models, fine-tuned from the same pre-trained model through weight interpolation, often leads to a more balanced performance among the original models~\citep{ilharco2023editing,yadav2023tiesmerging,yu2024language}.
Motivated by this insight, we conducted initial investigations into merging an RLHF model with the reference SFT model it trained from.
Our observations indicate that this offline model merging effectively mitigates the alignment cost. 
As depicted in \Cref{tab:main_results}, the offline merged model restores performance comparable to the SFT model across language benchmarks and language proficiency. 
However, this improvement comes at the expense of a reduction in preference rewards compared to the RLHF model.

The modest performance gains from offline merging are unsurprising, given that single-time parameter interpolation only allows for trade-offs between models with fixed capabilities.
During the RLHF training process, each optimization step improves the model's capabilities. Thus, we have the chance to ensure that the direction of these changes aligns with the reference SFT model. 
In this paper, we integrate model merging into each RLHF optimization step and introduce the Online Merging Optimizer.
This innovative optimizer enhances rewards more effectively than traditional optimizers like AdamW~\citep{loshchilov2018fixing} while simultaneously reducing the alignment tax, akin to offline merging. 
Specifically, we consider the alterations in model parameters before and after the SFT phase, termed delta parameters, as the final update direction for the SFT model. Throughout each RLHF step, we blend gradients with the SFT delta parameters, efficiently guiding the gradient to maximize rewards in the direction of SFT optimization. This results in well-calibrated gradients that balance reward maximization and alignment tax.

We conduct extensive experiments to demonstrate that our optimizer is effective across various dimensions: 
(1) for different backbones, including Qwen1.5~\citep{bai2023qwen} 
and LLaMA3~\footnote{\url{https://ai.meta.com/blog/meta-llama-3/}}; 
(2) for models of different sizes, ranging from 1.5B to 8B parameters;
(3) for various RLHF algorithms, such as DPO~\citep{rafailov2024direct}, IPO~\citep{amini2024direct}, and KTO~\citep{ethayarajh2024kto}; 
and (4) compatible with different offline merging algorithms, such as DARE~\citep{yu2024language} and TIES~\citep{,yadav2023tiesmerging}. 
Notably, the OnDARE optimizer consistently achieves better performance across 12 benchmarks in 7 categories along with MT-Bench and AlpacaEval 2.0, surpassing all regularization and offline merging baselines, demonstrating its superior capability as an optimizer designed for RLHF.

\section{Related Works}

\xhdr{Reinforcement Learning with Human Feedback~(RLHF)}
The objective of RLHF is to enhance the alignment of LLMs with human preferences and values, aiming to generate responses that are more helpful, accurate, and safer~\citep{ouyang2022training,Bai2022ConstitutionalAH,ji2023ai}.
Essentially, RLHF techniques treat human preferences as rewards and employ online or offline RL algorithms~\citep{zhao2023survey,rafailov2024direct,amini2024direct,ethayarajh2024kto,song2024preference,yuan2023rrhf}, to achieve this alignment. Typically, RLHF is built upon models trained via SFT~\citep{wang2023aligning,xu2024dpo,lu2024instag,wu2023selfevolved}.
One of the biggest challenges in RLHF is the bonus-tax trade-off~\citep{ouyang2022training,lin2024mitigating}, which shows RLHF models forget basic language modeling while pursuing high rewards.
In this work, we mitigate this issue by using online merging optimizers, which have proven effective across various sizes and backbone LLM families. 

\xhdr{Continual Learning~(CL)}
The Alignment Tax can be viewed as a form of catastrophic forgetting in CL. Catastrophic forgetting refers to a dramatic degradation of performance on old tasks when learning new tasks\citep{goodfellow2013empirical,kirkpatrick2017overcoming,wang2024comprehensive}. 
For example, constraining the shift in output space between reference and policy models \citep{ouyang2022training,rafailov2023direct}) is a common strategy employed in continual learning~\citep{li2017learning,rebuffi2017icarl,wang2024comprehensive}.
However, traditional methods mitigating forgetting issues in CL usually require data for previous training tasks for experience replay, which may be impractical for current open- or closed-source LLMs, whose alignment data is in-house and already multi-task.
Although we mainly experiment with our online merging optimizers on LLM alignment, they can be applied to a wider range of CL scenarios in the future.

\xhdr{Robust Fine-tuning}
Overcoming alignment tax can be seen from another perspective, similar to overcoming generalization degradation, involving the research area of robust fine-tuning. 
Nonetheless, robust fine-tuning methods often necessitate supplementary forward and backward computations, rendering them inefficient for large-scale models~\citep{liang2021rdrop,aghajanyan2021better,Jiang_2020,yuan2023hype}.
Some works aim to enhance robustness by eliminating invaluable parameter updates during training to achieve robust generalization~\citep{lee2020mixout,jiang2022rose,xu2021raise}.
Our proposed optimizer can be viewed as an extension of these gradient dropout methods. However, we further derive that merging the gradients with the delta parameters of the SFT model can better balance alignment rewards and tax. 
Comprehensive experiments show our methods surpass established robust fine-tuning baselines in alignment scenarios.

\xhdr{Offline Model Merging}
Model merging is widely studied as an effective training-free method to combine the capabilities of fine-tuned large language models~\citep{wortsman2022model,lin2023spurious,matena2022merging}. \citet{ilharco2023editing} proposed the concept of a "task vector" (also referred to as ``delta parameters'') and demonstrates that the arithmetic combination of them can adjust model behaviors. \citet{yadav2023tiesmerging} further addressed conflicts between different task vectors through norm-based sparsification and consensus on the signs of weights.
\citet{tam2024merging} employed the conjugate gradient method to find the optimal merging combination.
\citet{yu2024language} shown LLMs are highly robust to sparsify task vectors and proposes DARE for merging LLMs with random sparsification. 
Recently, \citet{lin2024mitigating} demonstrates that simple offline model averaging can help mitigate alignment tax.
Unlike these offline model merging methods that combine the parameters of different models in one shot, our work introduces an approach where, at each step of RLHF optimization, we online merge the gradients of the policy model with the delta parameters of the SFT model. 
Our results show that this optimizer is more effective than offline merging in boosting alignment rewards while eliminating alignment tax.


\begin{figure}
    \centering
    \includegraphics[width=\linewidth]{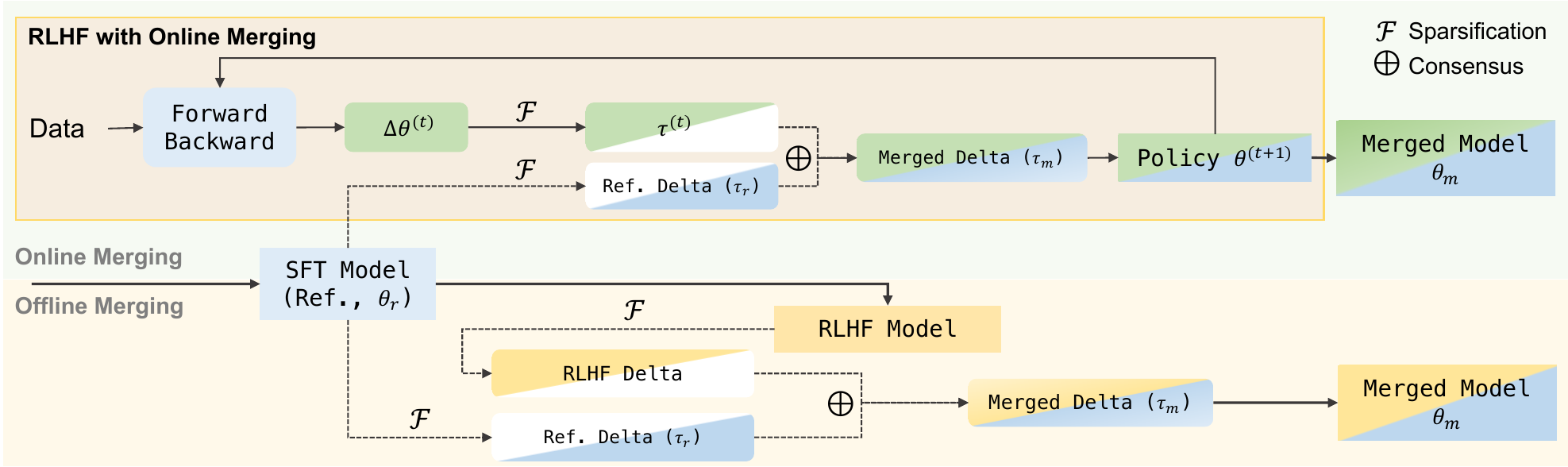}
    \caption{An illustrastion of RLHF with online merging optimizers described in \Cref{sec:methods}.
    In each RLHF iteration, we first obtain the update weight $\Delta\theta^{(t)}$, and then sparcify it and make a consensus with the delta parameters of the reference model.
    We use this merged delta as the update of the policy model in this iteration.
    We also compare online merging with offline merging, shown in the lower part of the figure and further introduced in \Cref{sec:offline_merging}.
    }
    \label{fig:main}
    \vspace{-2em}
\end{figure}


\section{Preliminaries and Motivations}\label{sec:offline_merging}

In this section, we begin by briefly defining alignment tax and introducing offline model merging. We then present preliminary results indicating that while offline merging can reduce alignment tax, it also diminishes rewards. These findings motivate our proposal for online merging optimizers.

\xhdr{Alignment Tax} Typically, aligning LLMs with human preferences involves two phases: SFT to establish an instruction-following model, followed by RLHF for enhanced human preference~\citep{ouyang2022training}. 
Current mainstream RLHF methods, such as PPO~\citep{ouyang2022training} and DPO ~\citep{rafailov2023direct}, guide the model to optimize the reward while incorporating a Kullback-Leibler (KL) divergence penalty between the output of learned RL policy and the reference SFT model. 
This penalty prevents the policy from deviating too far from its original goal of maintaining acquired language proficiency while pursuing preference rewards.
However, despite these efforts, RLHF models still exhibit fluctuations in performance across various NLP tasks and language capabilities~\citep{park2024disentangling}. 
We experimented with different KL divergence weights $\beta$ in the DPO setting, as shown in \Cref{fig:analysis_kl_constraint}. 
An increase in $\beta$ correlates with an elevation in the mean benchmark performance, albeit at the expense of diminished performance on MT-Bench and AlpacaEval.
Conversely, reducing $\beta$ leads the model to lose its fundamental abilities. 
Struggling to balance reward optimization and maintaining linguistic taxonomies has become the foremost challenge in RLHF training.

\xhdr{Model Merging} Instead of relying solely on the KL penalty to constrain the variation of the model's output space, we aim to find a trade-off path between the parameters of SFT and RLHF. 
This enables us to freely adjust whether the final model leans towards maximizing reward or minimizing tax.
Our objective relies on the discovery of Mode Connectivity~\citep{garipov2018loss}, which suggests that the local optima of modern deep neural networks are connected by simple curves, like a polygonal chain with only one bend. 
These connections are often categorized under model merging methods.
In this work, we specifically focus on general task arithmetic merging methods, which have proven effective in combining the abilities of fine-tuned models and have good characteristics such as linearity.
Here, we present a straightforward merging case where parameters $\theta_s$ and $\theta_r$, trained from the same pre-trained model $\theta_b$, are merged. The new capabilities learned by $\theta_s$ and $\theta_r$ relative to $\theta_b$ can be represented by their delta parameters: $\tau_s = \theta_s - \theta_b$ and $\tau_r = \theta_r - \theta_b$.
Merging methods typically begin by applying a sparsification operation to each delta parameter and then linearly combining them using consensus methods to derive the merged model $\theta_m = \theta_b + \tau_m = \theta_b + \mathcal{F}(\tau_s) \oplus \mathcal{F}(\tau_r)$, where $\mathcal{F}(\cdot)$ denotes a sparsification operator with corresponding hyper-parameters $p$, and $\oplus$ represents the consensus method used to handle parameter interference.
For instance, \citet{yadav2023tiesmerging} employed top-k sparsification and sign-based consensus, while \citet{yu2024language} utilized random sparsification and linear combination to achieve this merging process.

\xhdr{Offline Merging Mitigates Alignment Tax at Cost} To investigate whether Model Merging can restore the ability of the aligned model to recover the SFT model's capability, we conducted preliminary experiments. 
As shown in \Cref{tab:main_results}, we observe that the current mainstream model merging methods, when merging the SFT model and RLHF model, indeed improve performance across various linguistic abilities compared to the RLHF model, and outperform the SFT model on average. 
However, this improvement comes at the cost of decreased human preference, evidenced by a significant drop in MT-Bench and AlpcaEval 2.0 scores. 
In other words, while merging the SFT and RLHF models once can alleviate alignment tax, it also substantially diminishes the alignment bonus.

\xhdr{From Offline Merging to Online} It is not surprising that simply merging the SFT and RLHF models in an offline, one-time manner did not yield satisfactory results, as our reliance on the Mode Connectivity assumption has indicated that interpolating parameters results in interpolated model performance. 
Thus, finding a single point along this continuum that outperforms both ends of the spectrum simultaneously is indeed challenging.
However, through iterative model merging at each step of RLHF training, integrating the strengths of SFT with each update to maximize reward, we can progressively steer our training trajectory. 
This online merging could lead us to discover local minima that effectively balance foundational capabilities and human preferences.

\section{Online Merging Optimizers}\label{sec:methods}

Motivated by the offline merging, we explore incorporating model merging into the RLHF optimization steps in this section.
We begin by examining the commonly used gradient-based optimizers. 
At each optimization step $t$, LLM parameters are adjusted according to $\theta^{(t+1)} = \theta^{(t)} + \Delta\theta^{(t)}$, where $\Delta\theta^{(t)}$ is a gradient-based weight modification.
For instance, in the Adam optimizer~\citep{kingma2017adam}, $\Delta\theta$ is computed using the normalized gradient by exponential moving average, as shown in \Cref{alg:online_merging_optimizer}.
From the perspective of model merging, in RLHF training, $\Delta\theta^{(t)}$ can be seen as a representation of improved human preference—a set of parameters ready to be merged into the policy model at step $t$.
By post-processing $\Delta\theta^{(t)}$ to encapsulate not only human preference but also the basic capabilities of the SFT model, we can achieve an online merging of the RLHF policy model and the SFT model. 
To implement this concept, we propose an online merging optimizer. 

\subsection{From Gradient-based Optimizer to Online Merging Optimizer}

As introduced in \Cref{sec:offline_merging}, offline task arithmetic techniques merge LLMs by aggregating their delta parameters as $\theta_m = \theta_b + \tau_m=\theta_b+\mathcal{F}(\tau_s) \oplus \mathcal{F}(\tau_r)$.
In alignment, we aim to merge the reference SFT model $\theta_r$ and the policy model $\theta^{(t+1)}$ at the $t$-th training step.
Here, we define the merged parameter at step $t$ as:

\begin{align}
    \theta^{(t+1)} &= \theta_b + \mathcal{F}(\theta^{(t)} - \theta_b + \Delta\theta^{(t)}) \oplus \mathcal{F}(\theta_r - \theta_b) = \theta_b + \mathcal{F}(\tau^{(t)} + \Delta\theta^{(t)}) \oplus \mathcal{F}(\tau_r),\label{eq:estimation_begin}
\end{align}

where $\theta_b$ is the the parameters of pre-trained model that the reference and policy models are finetuned on, and $\tau_r, \tau^{(t)}$ are delta parameters $\tau_r=\theta_r-\theta_b$, $\tau^{(t)} = \theta^{(t)} - \theta_b$.

However, we empirically find directly optimizing \Cref{eq:estimation_begin} is unstable and hard to converge, and \Cref{eq:estimation_begin} requires an additional cache for pre-trained model parameters.
Therefore, we apply a relaxation on \Cref{eq:estimation_begin} by moving the $\tau^{(t)}$ out of the merging and only focus on $\Delta\theta^{(t)}$, meaning we relax the merging for the past but only keep it toward the future update: 
\begin{align}
    \theta^{(t+1)} \sim \theta_b + \tau^{(t)} + \mathcal{F}(\Delta\theta^{(t)}) \oplus \mathcal{F}(\tau_r)
    = \theta^{(t)} + \mathcal{F}(\Delta\theta^{(t)}) \oplus \mathcal{F}(\tau_r).\label{eq:estimation_end}
\end{align}
The delta parameter $\tau^{(t)}$ is already sparsified and consolidated with the reference in the previous update, so this relaxation still aligns well with our motivation for applying offline merging in optimization steps.
Another essential benefit of this relaxation is avoiding caching additional parameters of $\theta_b$, enhancing memory efficiency.
With this relaxation, we show that the online merging at each optimization step can be approximated by consolidation between gradient-based delta weights and the delta parameters of the reference model.

\subsection{Implementations}
Note that our optimizer framework is highly flexible and compatible with exiting model merging methods.
In this work, we develop two online merging optimizers, OnDARE and OnTIES, based on widely-used model merging methods, \textsc{DARE}~\citep{yu2024language} and \textsc{TIES}~\citep{yadav2023tiesmerging}.

\xhdr{OnDARE Optimizer} DARE employs a random sparsification method and a linear combination as the consensus method. So, the OnDARE optimizer following \Cref{eq:estimation_end} is displayed in \Cref{eq:ondare}:
\begin{equation}\label{eq:ondare}
  \theta_m^{(t)} = \theta_m^{(t-1)} + (1-\alpha)\mathcal{F}_{R}(\Delta\theta) +\alpha\mathcal{F}_{R}(\tau_r),
\end{equation}
where $\alpha$ is a hyper-parameter controlling the weight of delta parameters of the reference model.
Larger $\alpha$ introduces stronger regularization on the training.
$F_R(\cdot)$ shown in \Cref{eq:random-spar} is a random sparsification operator based on the Bernoulli distribution with a fixed reserving probability $p$:
\begin{equation}\label{eq:random-spar}
\mathcal{F}_{R}(x)_i = \left\{
\begin{array}{ll}
x_i,&\mathrm{Bernoulli}(p) = 1\\
0,&\mathrm{Bernoulli}(p) = 0\\
\end{array}
\right.
\end{equation}


\xhdr{OnTIES Optimizer}
TIES uses a top-k percentage sparsification and a sign-based consensus method.
Specifically, it reserves the top-p percentage of parameters regarding absolute values from each merging candidate. It calculates the element-wise majority signs based on the sign and norms, drops parameters with different signs with the majority, and weighted sum up the rest.
Similarly, the formulation of the OnTIES optimizer is shown in \Cref{eq:onties}:
\begin{equation}\label{eq:onties}
    \theta_m^{(t)} = \theta_m^{(t-1)} + (1-\alpha)\mathcal{F}_{R}(\Delta\theta) \oplus_{sign}\alpha\mathcal{F}_{R}(\tau_r),
\end{equation}
where $\mathcal{F}_{R}(\cdot)$ is a top-p percentage sparsification detailed in \Cref{eq:top-p-spar}.
The consensus method $\oplus_{sign}$ used in the OnTIES optimizer is a sign-based consensus among sparsified delta parameters as shown in \Cref{eq:sign-consensus}, where $\mathbb{I}$ is the indicator function:
\begin{align}\label{eq:top-p-spar}
\mathcal{F}_{R}(x)_i = \left\{
\begin{array}{ll}
x_i, & \vert x_i \vert \in \textrm{top-p}(\vert x\vert)\\
0, & o.w.\\
\end{array}
\right.
\end{align}
\begin{align}\label{eq:sign-consensus}
a\oplus_{sign} b = \left\{
\begin{array}{ll}
a+b, & \textrm{sign}(a) = \textrm{sign}(b)\\
a\mathbb{I}(\vert a \vert \ge \vert b \vert) + b\mathbb{I}(\vert b \vert \ge \vert a \vert), & \textrm{sign}(a) \neq \textrm{sign}(b)\\
\end{array}
\right.
\end{align}

The pseudo-code examples for these variants of online merging optimizers are shown in \Cref{alg:online_merging_optimizer}.
It is worth noticing that we neglect the rescaling of sparsified delta parameters, which are proven to be essential in the offline merging~\citep{yu2024language,yadav2023tiesmerging}, as we find rescaling harms the numeric stabilities in multi-step optimization.
\section{Experiments}

\subsection{Experimental Setup}\label{sec:experimental_setup}

\textbf{Dataset.}
We conduct our experiment on the widely-used preference dataset \textsc{UltraFeedback}.
To be more specific, we use the binarized version of \textsc{UltraFeedback} from \citet{tunstall2023zephyr}\footnote{\url{https://huggingface.co/datasets/HuggingFaceH4/ultrafeedback_binarized}}, which is a reannotated and clean dataset for preference learning, compared with the original release~\citep{cui2023ultrafeedback}.
The training and evaluation splits of \textsc{UltraFeedback} contain about 61K and 2K preference pairs, respectively, ranked by GPT-4, cleaned with manual efforts, and decontaminated with popular benchmarks such as TruthfulQA. The prompts in \textsc{UltraFeedback} are large-scale, fine-grained, and diverse from multiple sources.

\textbf{Training.}
We mainly explore our online merging optimizers in direct preference optimization~(DPO, \cite{rafailov2023direct}) on the \textsc{UltraFeedback} dataset, as it is widely applied in nowadays large-scale LLM alignment due to its lower cost of training compared with Proximal Policy Optimization~(PPO, \citep{schulman2017proximal}).
The general DPO includes sampling and annotating the responses from the policy model.
In this work, we use an off-policy setting of DPO that directly trained our policy models on \textsc{UltraFeedback} dataset, which has also proved effective in enhancing the helpfulness and harmlessness.
The training loss of DPO is shown in \Cref{eq:dpo}, where $\beta$ is the hyper-parameter controlling the strength of KL penalty between policy and reference models, $\pi_{\theta}$ and $\pi_{\textrm{ref}}$ are policy and reference models, $y_w$ and $y_l$ are chosen and rejected samples in the preference pairs, respectively:
\begin{equation}\label{eq:dpo}
    \mathcal{L}(\pi_{\theta},\pi_{\textrm{ref}}) = -\mathbb{E}_{x,y_w,y_l\sim D}[\log\sigma(\beta\log\frac{\pi_{\theta}(y_w\vert x)}{\pi_{\textrm{ref}}(y_w\vert x)} - \beta\log\frac{\pi_{\theta}(y_l\vert x)}{\pi_{\textrm{ref}}(y_l\vert x)})].
\end{equation}
Note that online merging optimizers are agnostic with the training loss and optimizers, so they can be easily applied to PPO,  other variants of DPO~\citep{azar2023general,hong2024orpo,ethayarajh2024kto}, and other online preference learning methods~\citep{wu2024selfplay,guo2024direct}.

\textbf{Evaluation.}
Assessing aligned large language models is a challenging task.
We follow the straightforward principle that comprehensive evaluation yields trustworthy assessment.
Therefore, our evaluation includes 12 public benchmarks in 7 categories evaluating the comprehensive abilities of aligned LLMs:
\textbf{Mathematics}:
    (1) GSM8K~\citep{cobbe2021training}
    (2) Math401~\citep{yuan2023large}
    (3) Math23K~\citep{wang-etal-2017-deep};
\textbf{Coding}:
    (1) HumanEval~\citep{chen2021codex}
    (2) MBPP~\citep{austin2021program}
    (3) DS1000~\citep{lai2022ds1000}
    (4) CodeApex~\citep{fu2024codeapex};
\textbf{Instruction-following~(IF)}:
    (1) IFEval~\citep{zhou2023instructionfollowing};
\textbf{Reading Comprehension~(RC)}:
    (1) COQA~\citep{reddy2019coqa}
    (2) DROP~\citep{dua2019drop};
\textbf{Knowledge}
        (1) MMLU~\citep{hendrycks2021measuring};
\textbf{Agent}
        (1) NousResearch;
\textbf{Code-switching};
We report the details of benchmarks in \Cref{app:benchmark_details} and inference configurations in \Cref{app:configuration}.
We use average scores within a category as the final category score and an average score among all benchmarks in all categories as the overall assessment.
We also involve MT-Bench~\footnote{\url{https://huggingface.co/spaces/lmsys/mt-bench}} and AlpacaEval 2.0~\citep{dubois2023alpacafarm,dubois2024length} with length-controlled scoring, two leading and popular benchmarks that assess LLMs' alignment with human preferences using GPT-4-based evaluators.

\xhdr{Baselines}
A naive baseline of our methods is the vanilla AdamW. We further consider offline merging methods, such as linear merging, DARE~\cite{yu2024language}, and TIES~\citep{yadav2023tiesmerging} as our strong baselines as \citet{lin2024mitigating} shows simple merging can mitigate alignment tax.
As alignment tax is related to forgetting, we also involve traditional regularization methods, such as the KL penalty~\citep{ouyang2022training}, EMA~\citep{hunter1986exponentially,noukhovitch2023language}, and ChildTuning~\citep{xu2021raise} as our baselines.
Specifically, the DPO algorithm adjusts the KL penalty with the hyper-parameter $\beta$.
We also include LoRA~\citep{hu2021lora} as one of our baselines, as parameter-efficient methods apply regularization regarding the weight space in the training~\citep{biderman2024lora}.
Implementations of baselines are detailed in \Cref{app:configuration}.

\xhdr{Configurations}
We experiment with three LLM sizes, Qwen1.5-1.8B, Qwen1.5-7B, and LLaMa-3-8B series.
Specifically, we use Qwen-1.8B-Base, Qwen-7B-Base, and LLaMa-3-8B as the base models for online optimizers.
And we conduct direct preference optimization on \textsc{UltraFeedback} on Qwen1.5-1.8B-SFT, Qwen1.5-7B-SFT, and LLaMa-3-8B-it as reference models.
Two Qwen1.5 supervised finetuned models were trained on multi-lingual instruction data but no overlap with the \textsc{UltraFeedback} dataset.
More configuration about hyper-parameter searching, statistical significance, and experiment resources are detailed in \Cref{app:configuration}

\definecolor{lightgreen}{RGB}{229, 255, 229}
\definecolor{lightred}{RGB}{255, 229, 229}

\begin{table}[!t]
    \centering
    \caption{
        Main results of baselines and online merging optimizers on \textsc{UltraFeedback} trained from various backbone LLMs.
        In merging and dropping settings, we experiment with two variants, OnDARE and OnTIES.
        I.F., R.C., and Know. are short for instruction-following, reading comprehension, and knowledge.
        The Avg. columns are average scores except for MT-Bench 
        and AlpacaEval 2.0.
        Benchmarks in each category are detailed in \Cref{sec:experimental_setup}.
        We report the improvement against the vanilla AdamW (marked in {\setlength{\fboxsep}{1pt}\colorbox{gray!40}{gray}}).
        Higher scores are marked in darker {\setlength{\fboxsep}{1pt}\colorbox{green!40}{green}}, while lower in darker {\setlength{\fboxsep}{1pt}\colorbox{red!40}{red}}.
        The best and second-best results are marked in \textbf{bold} and \underline{underline}, respectively.
        \\
    }
    \small
    \resizebox{.9\textwidth}{!}{
    \setlength{\tabcolsep}{2mm}{
    \begin{tabular}{p{2mm}l|cccccccc|cc}
    \toprule
    & \textbf{Method} & \textbf{Math} & \textbf{Code} & \textbf{I.F.} & \textbf{R.C.} & \textbf{Know.} & \textbf{Agent} & \textbf{CodeSwitch} & \textbf{Benchmark Avg.} & \textbf{MT-Bench} & \textbf{AlpacaEval 2.0 (LC)} \\
    \midrule
\multirow{14}{*}{\rotatebox[origin=c]{90}{\textbf{Qwen1.5-1.8B-Chat}}} & Reference & 37.3 & 7.8 & 23.1 & 52.2 & 43.1 & \underline{70.8} & 58.6 & 41.8 & 4.37 & 3.99 \\
\rowcolor[gray]{0.95}\cellcolor{white} & AdamW & \textbf{37.8} & 8.4 & 23.1 & 50.8 & 43.2 & 67.5 & 61.9 & 41.8 & 4.59 & 4.79 \\

 & \multicolumn{10}{c}{\rule{0pt}{10pt}\multirow{1}{*}[1pt]{\textit{Regularization}}}\\

 & KL Penalty & \cellcolor{lightred!13.0}{-\underline{0.3}} & 0.0 & 0.0 & \cellcolor{lightgreen!10.0}{+0.5} & \cellcolor{lightgreen!38.0}{+0.3} & \cellcolor{lightgreen!21.0}{+1.2} & \cellcolor{lightgreen!40.0}{+0.8} & \cellcolor{lightgreen!80.0}{+\underline{0.4}} & 0.00 & \cellcolor{lightgreen!100.0}{+0.02}\\
 & EMA & \cellcolor{lightred!83.0}{-2.0} & \cellcolor{lightred!83.0}{-0.5} & \cellcolor{lightred!100.0}{-0.8} & \cellcolor{lightred!100.0}{-1.4} & \cellcolor{lightgreen!13.0}{+0.1} & \cellcolor{lightgreen!54.0}{+3.0} & \cellcolor{lightgreen!80.0}{+\underline{1.6}} & 0.0 & \cellcolor{lightgreen!37.0}{+0.09} & \cellcolor{lightred!100.0}{-0.10} \\
 & ChildTuning & \cellcolor{lightred!38.0}{-0.9} & \cellcolor{lightgreen!50.0}{+\underline{0.1}} & \cellcolor{lightred!25.0}{-0.2} & \cellcolor{lightgreen!39.0}{+\underline{1.9}} & \cellcolor{lightgreen!88.0}{+\underline{0.7}} & 0.0 & \cellcolor{lightgreen!20.0}{+0.4} & \cellcolor{lightgreen!60.0}{+0.3} & \cellcolor{lightgreen!12.0}{+0.03} & \cellcolor{lightred!100.0}{-0.04}\\
 & LoRA & \cellcolor{lightred!54.0}{-1.3} & \cellcolor{lightgreen!100.0}{+\textbf{0.2}} & \cellcolor{lightgreen!73.0}{+0.8} & \cellcolor{lightgreen!33.0}{+1.6} & \cellcolor{lightgreen!38.0}{+0.3} & \cellcolor{lightgreen!5.0}{+0.3} & \cellcolor{lightred!100.0}{-2.5} & \cellcolor{lightred!100.0}{-0.1} & \cellcolor{lightred!70.0}{-0.07} & \cellcolor{lightred!100.0}{-0.23} \\

 & \multicolumn{10}{c}{\rule{0pt}{10pt}\multirow{1}{*}[1pt]{\textit{Offline Merging}}}\\

 & Linear & \cellcolor{lightred!71.0}{-1.7} & 0.0 & \cellcolor{lightgreen!73.0}{+0.8} & \cellcolor{lightred!21.0}{-0.3} & \cellcolor{lightred!33.0}{-0.2} & \cellcolor{lightgreen!54.0}{+3.0} & 0.0 & \cellcolor{lightgreen!40.0}{+0.2} & \cellcolor{lightred!30.0}{-0.03} & \cellcolor{lightred!100.0}{-0.96} \\
 & DARE & \cellcolor{lightred!100.0}{-2.4} & \cellcolor{lightred!50.0}{-0.3} & \cellcolor{lightgreen!18.0}{+0.2} & \cellcolor{lightred!86.0}{-1.2} & \cellcolor{lightred!17.0}{-0.1} & \cellcolor{lightgreen!100.0}{+\textbf{5.6}} & \cellcolor{lightred!32.0}{-0.8} & \cellcolor{lightgreen!40.0}{+0.2} & \cellcolor{lightgreen!37.0}{+0.09} & \cellcolor{lightred!100.0}{-0.72} \\
 & TIES & \cellcolor{lightred!67.0}{-1.6} & \cellcolor{lightred!33.0}{-0.2} & \cellcolor{lightgreen!73.0}{+0.8} & \cellcolor{lightgreen!4.0}{+0.2} & \cellcolor{lightred!100.0}{-0.6} & \cellcolor{lightgreen!27.0}{+1.5} & \cellcolor{lightred!32.0}{-0.8} & \cellcolor{lightred!100.0}{-0.1} & \cellcolor{lightred!100.0}{-0.10} & \cellcolor{lightred!100.0}{-0.41} \\

 & \multicolumn{10}{c}{\rule{0pt}{10pt}\multirow{1}{*}[1pt]{\textit{Online Merging~(Ours)}}}\\

 & OnTIES & \cellcolor{lightred!83.0}{-2.0} & \cellcolor{lightred!100.0}{-0.6} & \cellcolor{lightgreen!91.0}{+\underline{1.0}} & \cellcolor{lightred!36.0}{-0.5} & \cellcolor{lightgreen!25.0}{+0.2} & \cellcolor{lightgreen!27.0}{+1.5} & \cellcolor{lightgreen!100.0}{+\textbf{2.0}} & \cellcolor{lightgreen!60.0}{+0.3} & \cellcolor{lightgreen!62.0}{+\underline{0.15}} & \cellcolor{lightgreen!100.0}{+\textbf{0.12}} \\
 & OnDARE & \cellcolor{lightred!54.0}{-1.3} & \cellcolor{lightred!100.0}{-0.6} & \cellcolor{lightgreen!100.0}{+\textbf{1.1}} & \cellcolor{lightgreen!100.0}{+\textbf{4.9}} & \cellcolor{lightgreen!100.0}{+\textbf{0.8}} & \cellcolor{lightred!100.0}{-1.7} & \cellcolor{lightgreen!20.0}{+0.4} & \cellcolor{lightgreen!100.0}{+\textbf{0.5}} & \cellcolor{lightgreen!100.0}{+\textbf{0.24}} & \cellcolor{lightgreen!100.0}{+\underline{0.05}} \\
\midrule
\multirow{14}{*}{\rotatebox[origin=c]{90}{\textbf{Qwen1.5-7B-Chat}}} & Reference & 65.5 & 23.0 & 34.3 & 69.2 & 59.1 & 74.6 & 75.0 & 57.3 & 6.97 & 10.71 \\
\rowcolor[gray]{0.95}\cellcolor{white} & AdamW & 69.3 & 22.1 & 38.0 & 71.9 & 59.4 & 76.3 & 71.7 & 58.4 & \underline{7.18} & 11.66\\ 

 & \multicolumn{10}{c}{\rule{0pt}{10pt}\multirow{1}{*}[1pt]{\textit{Regularization}}}\\

 & KL Penalty & \cellcolor{lightred!68.0}{-1.9} & \cellcolor{lightgreen!47.0}{+0.9} & \cellcolor{lightred!92.0}{-2.2} & \cellcolor{lightred!67.0}{-2.0} & \cellcolor{lightgreen!10.0}{+0.1} & \cellcolor{lightgreen!84.0}{+\underline{3.2}} & \cellcolor{lightgreen!55.0}{+2.9} & \cellcolor{lightgreen!9.0}{+0.1} & \cellcolor{lightred!72.0}{-0.23} & \cellcolor{lightgreen!100.0}{+0.08} \\
 & EMA & \cellcolor{lightred!71.0}{-2.0} & \cellcolor{lightgreen!26.0}{+0.5} & \cellcolor{lightred!79.0}{-1.9} & \cellcolor{lightred!100.0}{-3.0} & \cellcolor{lightred!67.0}{-0.4} & \cellcolor{lightgreen!39.0}{+1.5} & \cellcolor{lightgreen!70.0}{+3.7} & \cellcolor{lightred!100.0}{-0.2} & \cellcolor{lightred!100.0}{-0.32} & \cellcolor{lightgreen!100.0}{+\textbf{1.24}}\\
 & ChildTuning & \cellcolor{lightred!100.0}{-2.8} & -0.47 & \cellcolor{lightgreen!53.0}{+1.0} & \cellcolor{lightred!88.0}{-2.1} & \cellcolor{lightred!37.0}{-1.1} & \cellcolor{lightred!100.0}{-0.6} & \cellcolor{lightgreen!68.0}{+2.6} & \cellcolor{lightgreen!47.0}{+2.5} & \cellcolor{lightred!50.0}{-0.1} & \cellcolor{lightred!34.0}{-0.11} \\
 & LoRA & \cellcolor{lightred!68.0}{-1.9} & \cellcolor{lightgreen!21.0}{+0.4} & \cellcolor{lightred!100.0}{-2.4} & \cellcolor{lightred!70.0}{-2.1} & \cellcolor{lightred!67.0}{-0.4} & \cellcolor{lightgreen!63.0}{+2.4} & \cellcolor{lightgreen!100.0}{+\textbf{5.3}} & \cellcolor{lightgreen!18.0}{+0.2} & \cellcolor{lightred!44.0}{-0.14} & \cellcolor{lightred!100.0}{-0.33} \\

 & \multicolumn{10}{c}{\rule{0pt}{10pt}\multirow{1}{*}[1pt]{\textit{Offline Merging}}}\\

 & Linear & \cellcolor{lightred!43.0}{-1.2} & \cellcolor{lightgreen!100.0}{+\textbf{1.9}} & \cellcolor{lightred!96.0}{-2.3} & \cellcolor{lightred!70.0}{-2.1} & \cellcolor{lightred!33.0}{-0.2} & \cellcolor{lightgreen!39.0}{+1.5} & \cellcolor{lightgreen!15.0}{+0.8} & \cellcolor{lightred!100.0}{-0.2} & \cellcolor{lightred!88.0}{-0.28} & \cellcolor{lightred!100.0}{-1.65} \\
 & DARE & \cellcolor{lightred!64.0}{-1.8} & \cellcolor{lightgreen!63.0}{+1.2} & \cellcolor{lightred!88.0}{-2.1} & \cellcolor{lightred!57.0}{-1.7} & \cellcolor{lightred!83.0}{-0.5} & \cellcolor{lightgreen!24.0}{+0.9} & \cellcolor{lightgreen!100.0}{+\textbf{5.3}} & \cellcolor{lightgreen!18.0}{+0.2} & \cellcolor{lightred!41.0}{-0.13} & \cellcolor{lightred!100.0}{-1.59} \\
 & TIES & \cellcolor{lightred!79.0}{-2.2} & \cellcolor{lightgreen!89.0}{+\underline{1.7}} & \cellcolor{lightred!8.0}{-0.2} & \cellcolor{lightred!47.0}{-1.4} & \cellcolor{lightred!17.0}{-0.1} & \cellcolor{lightgreen!100.0}{+\textbf{3.8}} & \cellcolor{lightgreen!47.0}{+2.5} & \cellcolor{lightgreen!55.0}{+\underline{0.6}} & \cellcolor{lightred!66.0}{-0.21} & \cellcolor{lightred!100.0}{-1.35} \\

 & \multicolumn{10}{c}{\rule{0pt}{10pt}\multirow{1}{*}[1pt]{\textit{Online Merging~(Ours)}}}\\

 & OnTIES & \cellcolor{lightred!79.0}{-2.2} & \cellcolor{lightgreen!5.0}{+0.1} & \cellcolor{lightgreen!22.0}{+0.5} & \cellcolor{lightgreen!25.0}{+0.3} & \cellcolor{lightgreen!100.0}{+\textbf{1.0}} & \cellcolor{lightgreen!63.0}{+2.4} & \cellcolor{lightgreen!25.0}{+1.3} & \cellcolor{lightgreen!45.0}{+0.5} & \cellcolor{lightred!25.0}{-0.08} & \cellcolor{lightgreen!100.0}{+\underline{0.91}} \\
  & OnDARE & \cellcolor{lightgreen!20.0}{+0.1} & \cellcolor{lightgreen!42.0}{+0.8} & \cellcolor{lightgreen!100.0}{+\textbf{2.3}} & \cellcolor{lightgreen!58.0}{+\underline{0.7}} & \cellcolor{lightgreen!30.0}{+0.3} & \cellcolor{lightred!39.0}{-0.9} & \cellcolor{lightgreen!77.0}{+4.1} & \cellcolor{lightgreen!100.0}{+\textbf{1.1}} & \cellcolor{lightgreen!60.0}{+\textbf{0.12}} & \cellcolor{lightgreen!100.0}{+0.28} \\
\midrule
\multirow{14}{*}{\rotatebox[origin=c]{90}{\textbf{LLaMa-3-8B-Instruct}}} & Reference & 67.8 & \underline{39.5} & 68.3 & 81.2 & 65.7 & 39.5 & 41.8 & 57.7 & 7.97 & 22.92 \\
 \rowcolor[gray]{0.95}\cellcolor{white} & AdamW & 68.3 & 39.3 & 68.4 & \underline{81.7} & 65.7 & 40.9 & 41.4 & 58.0 & 8.04 & 24.10 \\

 & \multicolumn{10}{c}{\rule{0pt}{10pt}\multirow{1}{*}[1pt]{\textit{Regularization}}}\\

 & KL Penalty & \cellcolor{lightgreen!100.0}{+\textbf{1.9}} & \cellcolor{lightred!62.0}{-0.8} & \cellcolor{lightred!100.0}{-0.6} & \cellcolor{lightred!27.0}{-0.4} & \cellcolor{lightgreen!100.0}{+\textbf{0.1}} & \cellcolor{lightgreen!29.0}{+1.2} & \cellcolor{lightgreen!73.0}{+5.7} & \cellcolor{lightgreen!77.0}{+\underline{1.0}} & \cellcolor{lightgreen!33.0}{+0.08} & \cellcolor{lightred!100.0}{-1.02} \\
 & EMA & \cellcolor{lightred!100.0}{-2.9} & \cellcolor{lightred!8.0}{-0.1} & \cellcolor{lightgreen!100.0}{+\textbf{1.0}} & \cellcolor{lightred!7.0}{-0.1} & \cellcolor{lightred!14.0}{-0.1} & \cellcolor{lightgreen!44.0}{+1.8} & \cellcolor{lightgreen!32.0}{+2.5} & \cellcolor{lightgreen!23.0}{+0.3} & \cellcolor{lightgreen!75.0}{+\underline{0.18}} & \cellcolor{lightgreen!100.0}{+0.34} \\
 & ChildTuning & \cellcolor{lightred!52.0}{-1.5} & \cellcolor{lightred!100.0}{-1.3} & \cellcolor{lightgreen!20.0}{+0.2} & \cellcolor{lightred!47.0}{-0.7} & \cellcolor{lightred!100.0}{-0.7} & \cellcolor{lightred!100.0}{-2.0} & \cellcolor{lightgreen!100.0}{+\textbf{7.8}} & \cellcolor{lightgreen!15.0}{+0.2} & \cellcolor{lightgreen!100.0}{+0.14} & \cellcolor{lightgreen!100.0}{+\underline{1.04}} \\
 & LoRA & \cellcolor{lightgreen!5.0}{+\underline{0.1}} & \cellcolor{lightred!46.0}{-0.6} & \cellcolor{lightred!33.0}{-0.2} & \cellcolor{lightred!27.0}{-0.4} & \cellcolor{lightgreen!100.0}{+\textbf{0.1}} & \cellcolor{lightgreen!100.0}{+\textbf{4.1}} & \cellcolor{lightgreen!21.0}{+1.6} & \cellcolor{lightgreen!46.0}{+0.6} & \cellcolor{lightgreen!29.0}{+0.07} & \cellcolor{lightred!100.0}{-0.16} \\

 & \multicolumn{10}{c}{\rule{0pt}{10pt}\multirow{1}{*}[1pt]{\textit{Offline Merging}}}\\

 & Linear & \cellcolor{lightred!38.0}{-1.1} & \cellcolor{lightred!15.0}{-0.2} & \cellcolor{lightgreen!30.0}{+0.3} & \cellcolor{lightred!80.0}{-1.2} & \cellcolor{lightred!71.0}{-0.5} & \cellcolor{lightred!40.0}{-0.8} & \cellcolor{lightred!100.0}{-4.5} & \cellcolor{lightred!100.0}{-1.2} & \cellcolor{lightgreen!46.0}{+0.11} & \cellcolor{lightgreen!100.0}{+0.02} \\
 & DARE & \cellcolor{lightred!34.0}{-1.0} & \cellcolor{lightred!23.0}{-0.3} & \cellcolor{lightred!17.0}{-0.1} & \cellcolor{lightred!13.0}{-0.2} & \cellcolor{lightred!57.0}{-0.4} & \cellcolor{lightgreen!37.0}{+1.5} & \cellcolor{lightgreen!58.0}{+4.5} & \cellcolor{lightgreen!38.0}{+0.5} & \cellcolor{lightgreen!58.0}{+0.14} & \cellcolor{lightgreen!100.0}{+0.18} \\
 & TIES & \cellcolor{lightred!41.0}{-1.2} & \cellcolor{lightgreen!100.0}{+\textbf{0.3}} & \cellcolor{lightred!50.0}{-0.3} & \cellcolor{lightred!100.0}{-1.5} & \cellcolor{lightred!43.0}{-0.3} & \cellcolor{lightgreen!44.0}{+1.8} & \cellcolor{lightgreen!15.0}{+1.2} & 0.0 & \cellcolor{lightgreen!4.0}{+0.01} & \cellcolor{lightgreen!100.0}{+0.21} \\

 & \multicolumn{10}{c}{\rule{0pt}{10pt}\multirow{1}{*}[1pt]{\textit{Online Merging~(Ours)}}}\\

 & OnTIES & \cellcolor{lightred!3.0}{-0.1} & \cellcolor{lightred!54.0}{-0.7} & 0.0 & \cellcolor{lightred!53.0}{-0.8} & \cellcolor{lightred!14.0}{-0.1} & \cellcolor{lightgreen!66.0}{+2.7} & \cellcolor{lightred!9.0}{-0.4} & 0.0 & \cellcolor{lightgreen!67.0}{+0.16} & \cellcolor{lightgreen!100.0}{+0.84} \\
  & OnDARE & \cellcolor{lightgreen!5.0}{+\underline{0.1}} & \cellcolor{lightred!100.0}{-1.3} & \cellcolor{lightgreen!80.0}{+\underline{0.8}} & \cellcolor{lightgreen!100.0}{+\textbf{0.5}} & \cellcolor{lightred!29.0}{-0.2} & \cellcolor{lightgreen!73.0}{+\underline{3.0}} & \cellcolor{lightgreen!85.0}{+\underline{6.6}} & \cellcolor{lightgreen!100.0}{+\textbf{1.3}} & \cellcolor{lightgreen!79.0}{+\textbf{0.19}} & \cellcolor{lightgreen!100.0}{+\textbf{1.57}} \\
    \bottomrule
    \end{tabular}}
    }
    \label{tab:main_results}
    \vspace{-2em}
\end{table}

\subsection{Main Results}

We present our main results in \Cref{tab:main_results}, which showcases the performance of baseline methods and our proposed online merging optimizers on \textsc{UltraFeedback}, trained from Qwen1.5-1.8B-Chat, Qwen1.5-7B-Chat, and LLaMa-3-8B-Chat.
Overall, regularization and offline model merging methods do not significantly improve the average performance of RLHF models on the benchmarks compared with vanilla AdamW in most settings, and lead to a decrease in preference scores on MT-Bench and AlpacaEval 2.0.
This indicates that simply relying on techniques such as dropout for gradients (ChildTuning), suppressing changes in model gradient updates (EMA), or making one-time adjustments to RLHF model parameters based on SFT models (Merging) are not effective solutions for the alignment bonus-tax trade-off.
The regularization baselines particularly work well for LLama-3-8B-It, as all regularization methods achieves consistent improvements on the average benchmark scores, as well as  MT-Bench and AlpacaEval scores.
In contrast, our proposed Online Merging Optimizer, particularly the OnDARE variant, achieved the most significant improvements on all the test sets.
OnDARE achieves the highest improvements on the benchmark average score and consistently enhances MT-Bench and AlpacaEval 2.0 on all three backbone LLMs, significantly surpasses other baselines, especially on the LLaMa-3-8B-Instruct experiments, where it achieves 1.3, 0.19, and 1.57 improvements on the benchmark, MT-Bench, and AlpacaEval, respectively.
Though OnTIES and OnDARE show effectiveness in boosting rewards and mitigating tax, OnDARE is slightly better than OnTIES in terms of the average benchmark scores in most cases, while OnTIES sometimes has higher LC win rates on AlpacaEval 2.0.
Detailed scores of each benchmark are reported in \Cref{tab:detail_results_llama3}, \Cref{tab:detail_results_qwen1.8}, \Cref{tab:detail_results_qwen7}.

\subsection{Hyper-parameter Effects}
This section analyzes how the two main hyper-parameters, parameter reserve rate $p$ and merging weight $\alpha$, affect the overall performance of online merging optimizers.

\textbf{Parameter Reserve Rate} is the reserve rate of parameters during online merging, which is $p$ in \Cref{eq:random-spar} and \Cref{eq:top-p-spar}.
We explore reserve rates ranging from $1$ to $1e^{-5}$ on Qwen1.5-1.8B-Chat to maximize the search space within our limited computational resources. As shown in \Cref{fig:analysis_param_reserve_rate}, online merging optimizers remain robust even at low parameter reserve rates down to $5e^{-4}$. This indicates that discarding 99.95\% of gradient-based parameter modifications at each RLHF step still results in stable training.
OnTIES is more sensitive to extremely low parameter reserve rates compared to OnDARE. This sensitivity arises because OnDARE employs an unbiased random sparsification method, whereas the top-k sparsification used by OnTIES introduces significant bias during training.


\begin{minipage}{0.5\textwidth}
    \centering
    \includegraphics[width=0.8\linewidth]{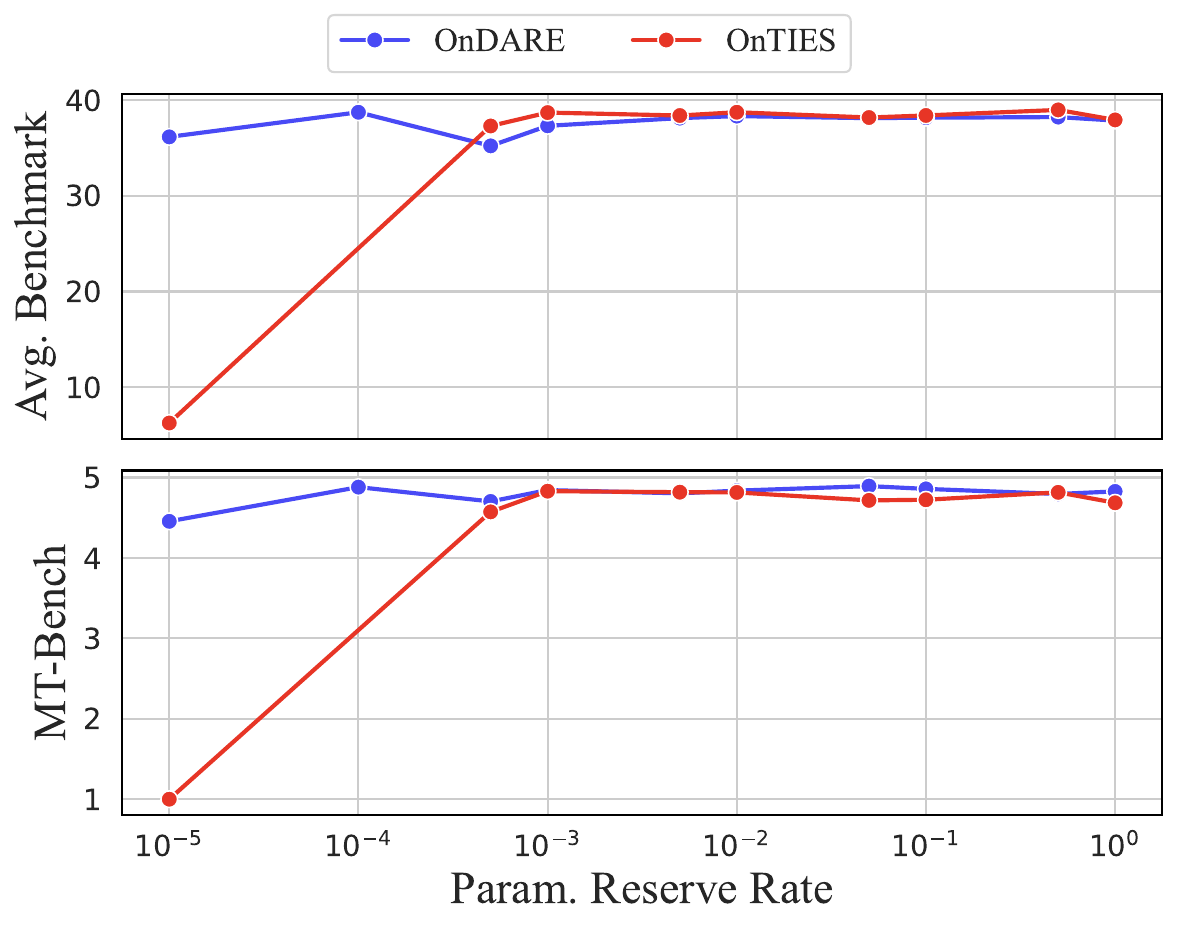}
    \captionof{figure}{Analysis of Parameter Reserve Rates}
    \label{fig:analysis_param_reserve_rate}
\end{minipage}%
\begin{minipage}{0.5\textwidth}
    \centering
    \captionof{table}{Analysis of the merging weight $\alpha$ with the OnDARE optimizer}
    \small
    \setlength{\tabcolsep}{3mm}{
    \begin{tabular}{ccc}
    \toprule
    \textbf{$\alpha$} & \textbf{Benchmarks(avg.)} & \textbf{MT-Bench} \\
    \midrule
    1e-4 & 7.4 & 1.00 \\
    5e-5 & 37.2 & 3.13 \\
    1e-5 & 40.1 & 4.37 \\
    5e-6 & \textbf{41.8} & 4.66 \\
    1e-6 & 41.6 & 4.76 \\
    5e-7 & 40.1 & 4.79 \\
    1e-7 & 39.2 & \textbf{4.85} \\
    \bottomrule
    \end{tabular}}
    \label{tab:analysis_merging_weights}
\end{minipage}

\textbf{Merging Weight} is the aggregated weight for gradients and delta parameters of the reference (SFT) model.
Larger merging weights introduce stronger regularization in the online merging optimizer.
We experiment various merging weights $\alpha$ from $10^{-4}$ to $10^{-7}$, and report the results at \Cref{tab:analysis_merging_weights}.
As the merging weight increases, the MT-Bench scores rise due to fewer regularizations being added during training, while the average benchmark scores initially increase but then decrease, peaking at $\alpha=5e^{-7}$.
Similar to the exponential coefficient in the EMA training~\citep{hunter1986exponentially}, large $\alpha$ leads to unstable training.
So, we suggest beginning the hyper-parameter searching of $\alpha$ from a minor number such as $10^{-7}$.
It is worth noticing that a special case $\alpha=0$ makes OnDARE equivalent to gradient dropout regularization methods, such as ChildTuning~\citep{xu2021raise} in our baselines.

\vspace{-1em}
\begin{wraptable}[16]{r}{0.45\textwidth}
    \vspace{-9pt}
    \centering
    \caption{Analysis on different RLHF algorithms with online merging optimizers. We report the average scores of benchmark and MT-Bench scores as \Cref{tab:main_results}.
    B.M. is short for Benchmark.
    Best scores are marked in \textbf{bold}.
    }
    \small
    \setlength{\tabcolsep}{0.8mm}{
    \begin{tabular}{clcc}
    \toprule
    \textbf{Method} & \textbf{Optimizer} & \textbf{B.M.(avg.)} & \textbf{MT-Bench} \\
     \midrule
    \multirow{3}{*}{IPO} & AdamW & \textbf{58.1} & 7.06 \\
     & OnDARE & 57.0 & 7.09 \\
     & OnTIES & 57.3 & \textbf{7.34} \\
     \midrule
    \multirow{3}{*}{KTO} & AdamW & 58.3 & 7.18 \\
     & OnDARE & 58.6 & \textbf{7.46} \\
     & OnTIES & \textbf{59.3} & 7.42 \\
    \bottomrule
    \end{tabular}}
    \label{tab:RLHF_method_ablation}
\end{wraptable}

\subsection{The Impact of RLHF Algorithms}\label{sec:rlhf_variants_analysis}

In \Cref{sec:experimental_setup}, we have validated the effectiveness of our Online Merging Optimizers in compatibility with DPO. In this section, we further study its application on other RLHF algorithms.
In this section, we further explore their application to other RLHF algorithms. Specifically, we implement OnDARE and OnTIES with IPO~\citep{azar2023general} and KTO~\citep{ethayarajh2024kto}. 
We train Qwen1.5-7B-Chat under these settings on the \textsc{UltraFeedback} datasets and present the results in \Cref{tab:RLHF_method_ablation}.
The Online Merging Optimizers, OnDARE, and OnTIES outperform AdamW across all algorithms on MT-Bench, except for the average benchmark score with the IPO algorithm. This demonstrates that their effectiveness extends to multiple RLHF algorithm variants.

\section{Discussion}


\subsection{Bridging Online and Offline Merging via Step-K Online Merging Optimizers}\label{sec:step_k_analysis}

Offline Model Merging occurs after RLHF training concludes, combining the trained model with the SFT model. In contrast, Online Model Merging merges models at each training step. 
To bridge the gap, we propose step-K online merging optimizers, as detailed in \Cref{alg:step-k_online_merging}.
In step-K online merging, the optimizer caches accumulated update $\Delta_c$, which is zero at the beginning of the training.
Then, it applies the standard Adam update to the parameters for $K$ optimization steps.
At the $K$-th step, it first reverts the parameters to their cached state from the previous $K$-th step and calculates the delta parameters as the difference between current and cached parameters.
These delta parameters are then merged with those of the reference model. 
The optimizer then updates the current parameters with the merged ones and refreshes the cached parameters.
The online gap step $K$ controls the \textbf{onlineness} for the step-K online merging optimizers.
The pseudo-codes are detailed in \Cref{alg:step-k_online_merging}.
When $K$ equals 1, the step-K online merging optimizer functions as the online merging optimizer introduced in \Cref{sec:methods}; When $K$ equals the total number of training steps, implying a single merge at the end of the training, the step-K online merging optimizer operates as the offline merging optimizer described in \Cref{sec:offline_merging}.

We then examine how the onlineness may affect the overall performance of step-K online merging optimizers.
We run step-K optimizers on \textsc{UltraFeedback} trained from Qwen1.5-1.8B-Chat with online gap steps in $\{5, 10, 50, 100, 200\}$, whose results are shown in \Cref{fig:analysis_online_step}.
As the online merging gap step increases, we witness a significant drop in MT-Bench scores for both OnDARE and OnTIES, with scores gradually approaching the performance of AdamW. 
The decline in average benchmark scores is less pronounced in MT-Bench, indicating that the "onlineness" of model merging is more critical for achieving higher rewards than for mitigating forgetting.

\begin{minipage}{0.5\textwidth}
    \centering
    \includegraphics[width=\linewidth]{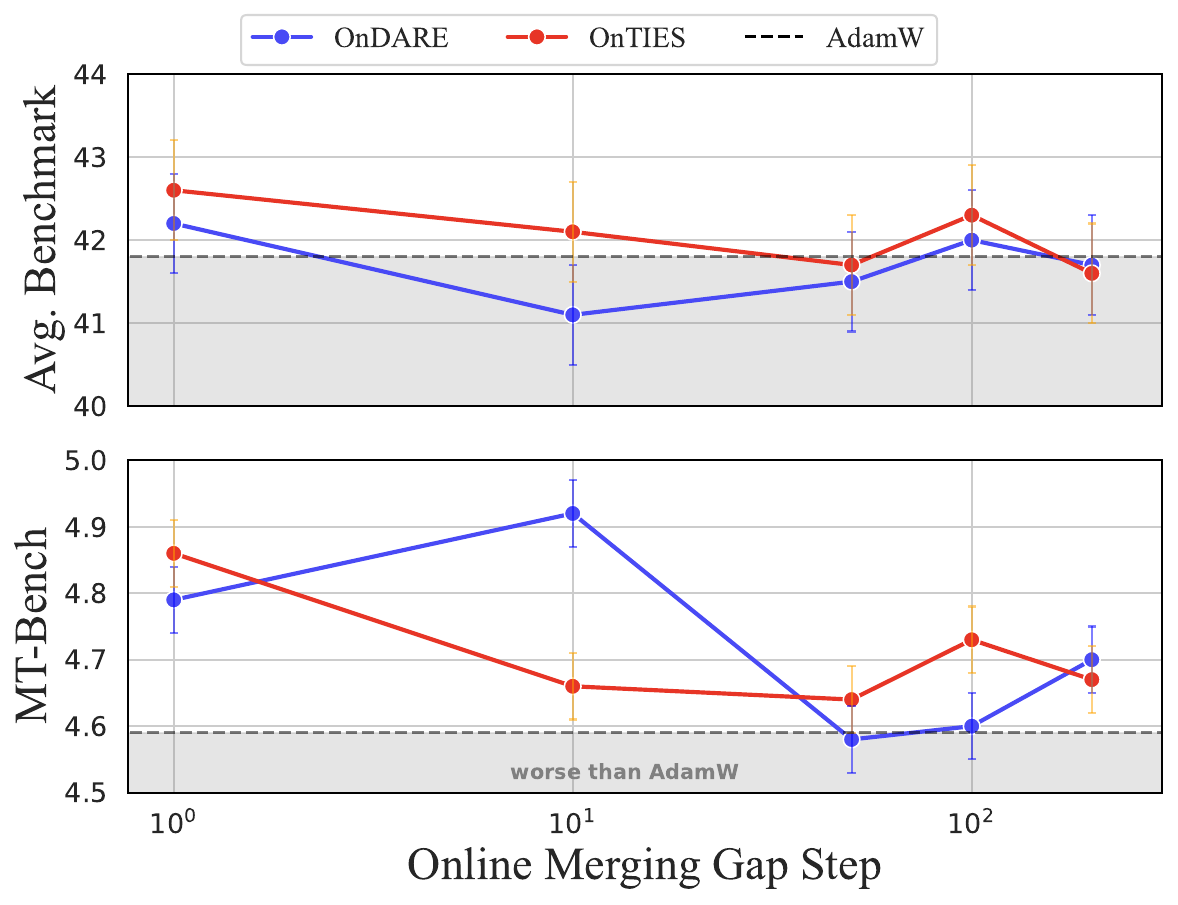}
    \captionof{figure}{Analysis of Online Merging Gap Step}
    \label{fig:analysis_online_step}
\end{minipage}
\begin{minipage}{0.5\textwidth}
    \centering
    \includegraphics[width=\linewidth]{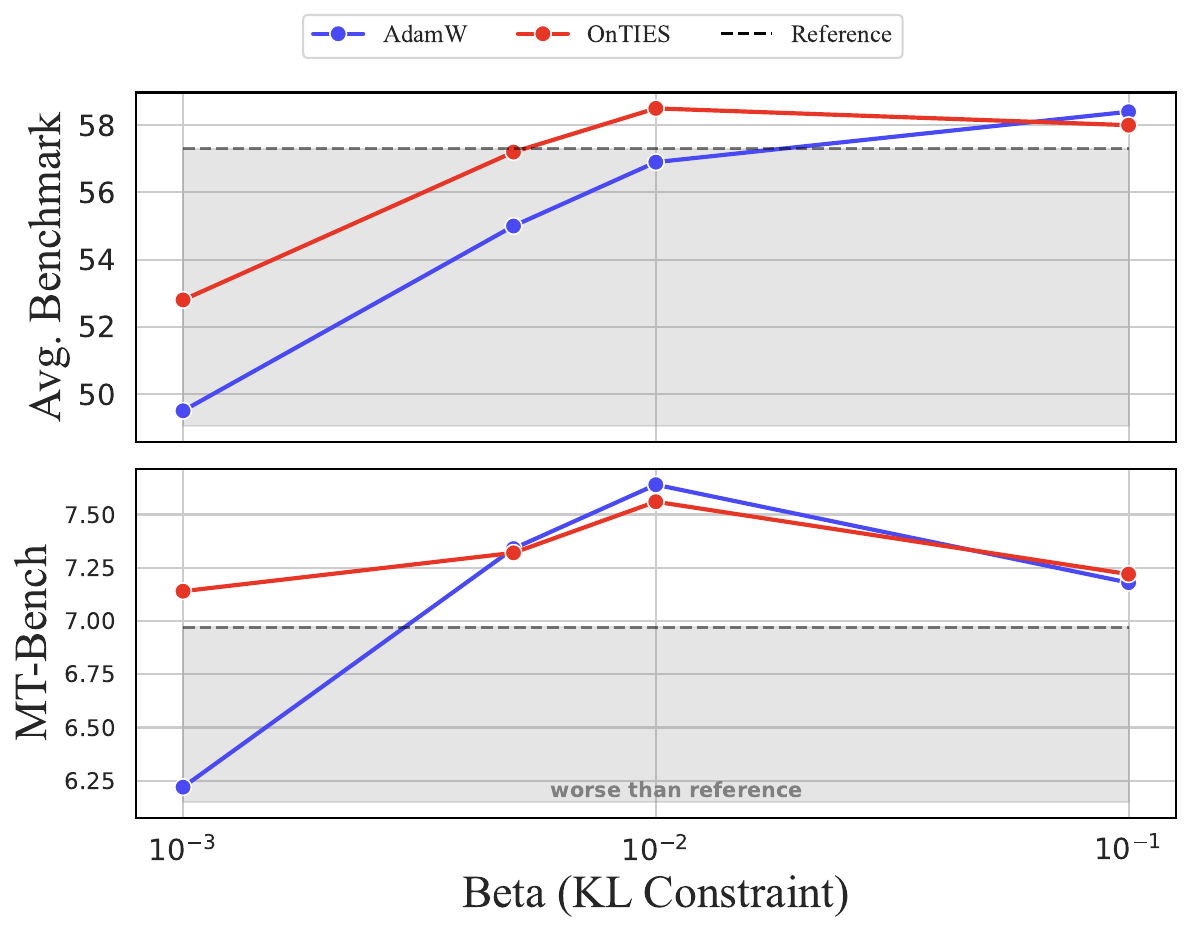}
    \captionof{figure}{Analysis of KL constraints}
    \label{fig:analysis_kl_constraint}
\end{minipage}

\subsection{Complementary Effect of KL Constraints and Online Merging}\label{sec:kl_analysis}

We further analyze the complementary effect of widely used KL constraint and online merging.
We conduct experiments on Qwen1.5-7B-Chat with AdamW and OnTIES under extremely small $\beta$ of the DPO algorithm, indicating very few KL constraints on the training.
The results are shown in \Cref{fig:analysis_kl_constraint}.
At $\beta=1e^{-1}$, AdamW and OnTIES show similar performance on both the MT-Bench and the average benchmark scores.
However, as the $\beta$ decreases under $1e^{-1}$, OnTIES is significantly better than AdamW on the average benchmark score, showing regularization from the online merging optimizer mitigates forgetting under lower KL constraints.
Furthermore, OnTIES still achieves MT-Bench over 6.9 (MT-Bench of the reference SFT model) under an extremely small $\beta$ ($1e^{-3}$), while AdamW falls to 6.25, demonstrating the complementary effect of KL constraints and online merging.
\section{Conclusion}\label{sec:conclusion}
In this work, we investigate the issue of alignment tax in the RLHF training of LLMs. 
Drawing inspiration from offline merging methods, we develop online merging optimizers that integrate general task arithmetic merging into each RLHF optimization step.
Extensive experiments conducted on various backbone LLMs demonstrate that online merging optimizers better mitigate alignment tax and achieve superior alignment performance compared to regularization and offline merging baselines. 
Additionally, we propose step-K online merging optimizers to bridge the gap between online and offline merging, providing in-depth analyses on the effects of hyper-parameters and ablations.
While our primary focus is on using online merging optimizers for LLM alignment, their applications can naturally extend to other domains facing catastrophic forgetting, such as the continual learning of LLMs. 
We hope that online merging optimizers will become a representative optimizer in future RLHF training, enabling researchers to more boldly pursue reward maximization without concerns about model degradation, ultimately producing more helpful, honest, and harmless LLMs.


\section*{Limitations}

Major limitations of online merging optimizers are related to the parameter efficiency.
Online merging optimizers enhance the memory requirements as they require a cache of additional delta parameters of the reference model as the counterpart of the delta updated weights in training.
At the same time, they can not be applied to the LoRA training, except for the cases where the reference models are also trained with LoRA adapters.
However, this limitation can be further eliminated by applying GaLore~\citep{zhao2024galore} with the online merging optimizers.

\section*{Acknowledgements}

We sincerely appreciate the support from Kai Dang, An Yang, Junyang Lin, and others on Qwen's Alignment team.
We thank Hongyi Yuan for providing insights on robust alignment and helping with engineering implementation, Le Yu for valuable feedback about model merging that improves our manuscript, and Tingyu Xia for visualization enhancement.

\bibliographystyle{acl_natbib}
\bibliography{ref.bib}
\clearpage

\appendix

\section*{Appendix}

\section{Detailed Algorithms}

\begin{algorithm}[H]
  \SetAlgoLined
  \SetKwInOut{Input}{Inputs}
  \SetKwInOut{States}{States}
  \Input{Reference and base models $\theta_r$, $\theta_b$, total step $T$, coefficients $\beta_1,\beta_2$, learning rate $\eta$, parameter reserve rate $p$, merging weight $\alpha$}
  \States{First momentum $m^{(t)}$, second momentum $v^{(t)}$ at step $t$, delta of the reference model $\tau_r$}

  Initialze $\tau_r=\theta_r - \theta_b$;
  
  \For{step $t$ $\le$ $T$}{
    \For{parameter $\theta^{(t-1)}$ in $\Theta^{(t-1)}$ and its gradient $g^{t}$}{
      \If{$t == 0$}{
        $m^{(0)} = \mathbf{0}$;
        $v^{(0)} = \mathbf{0}$;
    }
    $m^{(t)} = \beta_1 m^{(t-1)} + (1-\beta_1)g^{(t)}$;\\
    $v^{(t)} = \beta_2 v^{(t-1)} + (1-\beta_2)(g^{(t)})^2$;\\
    $\Delta\theta = -\eta m^{(t)} / \sqrt{v^{(t)}+\epsilon}$;\\
    \tcp{Online merging begin}
    \If{Using OnDARE}{
        $\theta^{(t)} = \theta^{(t-1)} + (1-\alpha)\mathcal{F}_{R}(\Delta\theta) +\alpha\mathcal{F}_{R}(\tau_r)$\\
        \tcp{Detailed in \Cref{eq:ondare},\ref{eq:random-spar}}
    }
    \ElseIf{Using OnTIES}{
        $\theta^{(t)} = \theta^{(t-1)} + (1-\alpha)\mathcal{F}_{R}(\Delta\theta) \oplus_{sign}\alpha\mathcal{F}_{R}(\tau_r)$\\
        \tcp{Detailed in \Cref{eq:onties},\ref{eq:top-p-spar},\ref{eq:sign-consensus}}
    }
    \tcp{Online merging end}
    }
  }
  \caption{Online Merging Optimizers~(OnDARE, OnTIES) based on Adam}
  \label{alg:online_merging_optimizer}
\end{algorithm}

\begin{algorithm}[htbp]
  \SetAlgoLined
  \SetKwInOut{Input}{Inputs}
  \SetKwInOut{States}{States}
  \Input{Reference and base models $\theta_r$, $\theta_b$, total step $T$, coefficients $\beta_1,\beta_2$, learning rate $\eta$, parameter reserve rate $p$, merging weight $\alpha$, \textbf{online merging gap step} $t_m$}
  \States{First momentum $m^{(t)}$, second momentum $v^{(t)}$ at step $t$, delta of the reference model $\tau_r$, \textbf{cached delta} $\Delta_c$}

  Initialze $\tau_r=\theta_r - \theta_b$;
  
  \For{step $t$ $\le$ $T$}{
    \For{parameter $\theta^{(t-1)}$ in $\Theta^{(t-1)}$ and its gradient $g^{t}$}{
      \If{$t == 0$}{
        $m^{(0)} = \mathbf{0}$;
        $v^{(0)} = \mathbf{0}$;
        $\Delta_c = \mathbf{0}$;
    }
    $m^{(t)} = \beta_1 m^{(t-1)} + (1-\beta_1)g^{(t)}$;\\
    $v^{(t)} = \beta_2 v^{(t-1)} + (1-\beta_2)(g^{(t)})^2$;\\
    $\Delta\theta = -\eta m^{(t)} / \sqrt{v^{(t)}+\epsilon}$;\\
    \If{current step \% $t_m$ == 0}{
        \tcp{Rollback current parameters}
        $p = p - \Delta_c$;\\
        $\Delta\theta = \Delta\theta + \Delta_c$;\\
        $\Delta_c = \mathbf{0}$;\\
        \If{Using OnDARE}{
            $\theta^{(t)} = \theta^{(t-1)} + (1-\alpha)\mathcal{F}_{R}(\Delta\theta) +\alpha\mathcal{F}_{R}(\tau_r)$\\
            \tcp{Detailed in \Cref{eq:ondare},\ref{eq:random-spar}}
        }
        \ElseIf{Using OnTIES}{
            $\theta^{(t)} = \theta^{(t-1)} + (1-\alpha)\mathcal{F}_{R}(\Delta\theta) \oplus_{sign}\alpha\mathcal{F}_{R}(\tau_r)$\\
            \tcp{Detailed in \Cref{eq:onties},\ref{eq:top-p-spar},\ref{eq:sign-consensus}}
        }
    }
    \Else{
        \tcp{Accumulate delta without merging}
        $\Delta_c = \Delta_c + \Delta\theta$;\\
        $\theta^{(t)} = \theta^{(t-1)} + \Delta\theta$;
    }
  }
  }
\caption{Step-K Online Merging Optimizers (OnDARE, OnTIES) based on Adam}
  \label{alg:step-k_online_merging}
\end{algorithm}

\section{Benchmark Details}\label{app:benchmark_details}

We all report the overall accuracy of the following benchmarks unless explaining with special annotations.
\begin{itemize}[leftmargin=1em]
    \setlength\itemsep{-0.1em}
    \item \textbf{Mathematics}:
        \begin{itemize}
            \item (1) GSM8K~\citep{cobbe2021training}: we use the full test set and test in the zero-shot setting with greedy decoding and report the accuracy;
            \item (2) Math401~\citep{yuan2023large}: we randomly downsample to obtain 101 samples and test in the zero-shot setting with default sampling decoding hyper-parameters and report the accuracy;
            \item (3) Math23K~\citep{wang-etal-2017-deep}: we randomly downsample to obtain 250 samples and we test in the zero-shot setting with default sampling decoding hyper-parameters and report the accuracy;
        \end{itemize}
    \item \textbf{Coding}:
        \begin{itemize}
            \item (1) HumanEval~\citep{chen2021codex}: we use the full test set and test in the zero-shot setting with greedy decoding and report the PASS@1;
            \item (2) MBPP~\citep{austin2021program}: we use the full test set, test in the zero-shot setting with default sampling decoding hyper-parameters, and report the PASS@1;
            \item (3) DS1000~\citep{lai2022ds1000}: we randomly downsample to obtain 600 samples, test in the zero-shot setting with default sampling decoding hyper-parameters, and report the PASS@1;
            \item (4) CodeApex~\citep{fu2024codeapex}: we use the full test set, test in the zero-shot setting with default sampling decoding hyper-parameters, and report the PASS@1;
        \end{itemize}
    \item \textbf{Instruction-following~(IF)}:
        (1) IFEval~\citep{zhou2023instructionfollowing}: we use the full test set, strictly follow the evaluation of IFEval, using greedy decoding and report strict instruction-level accuracy.
    \item \textbf{Reading Comprehension~(RC)}:
        \begin{itemize}
            \item (1) COQA~\citep{reddy2019coqa}: we use the full validation set, test in the zero-shot setting with default sampling decoding hyper-parameters, and report the recall of the keywords in the golden answer;
            \item (2) DROP~\citep{dua2019drop}: we use the full validation set (only the last turn in the session) and test in the zero-shot setting with default sampling decoding hyper-parameters, and report the recall of the keywords in the golden answer;
        \end{itemize}
    \item \textbf{Knowledge}
        (1) MMLU~\citep{hendrycks2021measuring}: we use the full test set and test in the zero-shot setting with greedy decoding;
    \item \textbf{Agent}
        (1) NousResearch: we report the right rates of function calling from four agent evaluation sets delivered by  NousResearch\footnote{\url{https://huggingface.co/datasets/NousResearch/json-mode-eval}\\\url{https://huggingface.co/datasets/NousResearch/func-calling-eval-singleturn}\\\url{https://huggingface.co/datasets/NousResearch/func-calling-eval-glaive}\\\url{https://huggingface.co/datasets/NousResearch/func-calling-eval}} with default decoding parameters.
    \item \textbf{Code-switching} We collect 113 queries for translation and 131 open-ended queries (most of which are in Chinese). For translation, we check if all tokens in the generated response are in the target language; For open-ended queries, we check if all tokens are in the same language of the query. The final score is the ratio of responses that do not have code-switching problem.
        
\end{itemize}

MT-Bench and AlpacaEval(LC) are judged by GPT-4 and GPT-4-1106-preview, respectively. MT-Bench has a standard deviation about 0.05, and AlpacaEval(LC) has a standard deviation about 0.5. Both are caused by the scoring instability of LLM-as-Judge.

\section{Configurations}\label{app:configuration}

\xhdr{Hyper-parameter Searching}
We use grid search for all optimal hyper-parameters in this work.
To be more specific, we search 6 learning rates \{$1e^{-5}$,$5e^{-5}$,$1e^{-6}$,$5e^{-6}$,$1e^{-7}$,$5e^{-7}$\} for all AdamW experiments to build strong baselines.
For hyper-parameters in baselines and proposed methods, we search at least three different values for each parameter to achieve optimal methods.
For every ablation study of online merging optimizers, we search 4 learning rates \{$5e^{-6}$,$1e^{-6}$,$1e^{-7}$,$5e^{-7}$\}.
We search batch-size in \{64, 128, 512\}, and we find batch-size 128 consistently works better for all settings with different backbone LLMs.
So, we run all main and ablation experiments with batch-size 128 without further annotations.

\xhdr{Generation Configuration}
We run models with their suggested generation hyper-parameters. We run the generation of LLaMa-3-8B-It with top-p 0.8 and temperature 0.7. We run the generation of the Qwen1.5 series with top-p 0.8, temperature 0.7, top-k 20, and repetition penalty 1.05. 

%

\xhdr{Baseline Implementations}
\begin{itemize}
    \item \textbf{AdamW}: We run AdamW with $\beta=0.1$ in DPO, which is also the same setting as our proposed online merging parameters.
    \item \textbf{KL Penalty}: We construct the KL Penalty baseline by searching the optimal $\beta$ from $1e^{-5}$ to $1$.
    \item \textbf{EMA}: We implement the exponential moving average in each optimization step and search the optimal exponential coefficient from $1e^{-2}$, $1e^{-3}$, and $1e^{-4}$. We find $1e^{-3}$ consistently achieves the best performance across all settings.
    \item \textbf{ChildTuning}: We implement the ChildTuning\textsubscript{F} following official implementation~\footnote{\url{https://github.com/RunxinXu/ChildTuning}}. We compare ChildTuning with our online merging optimizers under the same drop rates.
    \item \textbf{Offline Merging}: We use the implementation in \textit{mergekit}\footnote{\url{https://github.com/arcee-ai/mergekit}}. For each merging, we search three densities \{0.3, 0.5, 0.7\} and five weights \{0.1, 0.3, 0.5, 0.7, 0.9\} to make sure achieving the best baseline performance.
\end{itemize}

\xhdr{Experiment Resources}
We run all training on 64 NVIDIA H800 GPUs. Qwen1.5-1.8B-Chat, Qwen1.5-7B-Chat, and LLaMa-3-8B-it need training for about 3 hours.
We run all evaluations on 32 NVIDIA H800 GPUs for half an hour.

\xhdr{Asset Licences}
All Qwen models are distributed with the Tongyi Qwen Licence. All LLaMa-3 models used in this work are distributed with the LLaMa-3 Licence~\footnote{\url{https://llama.meta.com/llama3/license/}}.
The training dataset \textsc{UltraFeedback} and other open-source evaluation datasets are distributed with non-commercial licenses, such as the MIT Licence and Apache 2.0 Licence.

\section{Detailed Results}

\definecolor{lightgreen}{RGB}{229, 255, 229}
\definecolor{lightred}{RGB}{255, 229, 229}

\begin{table}[!t]
    \centering
    \caption{
        Main results of baselines and online merging optimizers on \textsc{UltraFeedback} trained from various backbone LLMs.
        In merging and dropping settings, we experiment with two variants, OnDARE and OnTIES.
        I.F., R.C., and Know. are short for instruction-following, reading comprehension, and knowledge.
        The Avg. columns are average scores except for MT-Bench and AlpacaEval.
        Benchmarks in each category are detailed in \Cref{sec:experimental_setup}.
        \\
    }
    \small
    \resizebox{.9\textwidth}{!}{
    \setlength{\tabcolsep}{1mm}{
    \begin{tabular}{p{2mm}l|cccccccc|cc}
    \toprule
    & \textbf{Method} & \textbf{Math} & \textbf{Code} & \textbf{I.F.} & \textbf{R.C.} & \textbf{Know.} & \textbf{Agent} & \textbf{CodeSwitch} & \textbf{Benchmark Avg.} & \textbf{MT-Bench} & \textbf{AlpacaEval(LC)} \\
    \midrule
\multirow{13}{*}{\rotatebox[origin=c]{90}{\underline{\textbf{Qwen1.5-1.8B-Chat}}}} & Reference & 37.3 & 7.8 & 23.1 & 52.2 & 43.1 & \underline{70.8} & 58.6 & 41.8 & 4.37 & 3.99\\
 & AdamW & \textbf{37.8} & 8.4 & 23.1 & 50.8 & 43.2 & 67.5 & 61.9 & 41.8 & 4.59 & 4.79\\

 & \multicolumn{11}{c}{\rule{0pt}{10pt}\multirow{1}{*}[1pt]{\textit{Regularization}}}\\

 & KL & \underline{37.5} & 8.4 & 23.1 & 51.3 & 43.5 & 68.7 & 62.7 & \underline{42.2} & 4.59 & 4.81 \\
 & EMA & 35.8 & 7.9 & 22.3 & 49.4 & 43.3 & 70.5 & \underline{63.5} & 41.8 & 4.68 & 4.69 \\
 & ChildTuning & 36.9 & \underline{8.5} & 22.9 & \underline{52.7} & \underline{43.9} & 67.5 & 62.3 & 42.1 & 4.62 & 4.75\\
 & LoRA & 36.5 & \textbf{8.6} & 23.9 & 52.4 & 43.5 & 67.8 & 59.4 & 41.7 & 4.52 & 4.56\\

 & \multicolumn{11}{c}{\rule{0pt}{10pt}\multirow{1}{*}[1pt]{\textit{Offline Merging}}}\\

 & linear & 36.1 & 8.4 & 23.9 & 50.5 & 43.0 & 70.5 & 61.9 & 42.0 & 4.56 & 3.83\\
 & DARE & 35.4 & 8.1 & 23.3 & 49.6 & 43.1 & \textbf{73.1} & 61.1 & 42.0 & 4.68 & 4.07 \\
 & TIES & 36.2 & 8.2 & 23.9 & 51.0 & 42.6 & 69.0 & 61.1 & 41.7 & 4.49 & 4.38\\

 & \multicolumn{11}{c}{\rule{0pt}{10pt}\multirow{1}{*}[1pt]{\textit{Online Merging~(Ours)}}}\\

 & OnTIES & 35.8 & 7.8 & \underline{24.1} & 50.3 & 43.4 & 69.0 & \textbf{63.9} & 42.1 & \underline{4.74} & 4.91 \\
 & OnDARE & 36.5 & 7.8 & \textbf{24.2} & \textbf{55.7} & \textbf{44.0} & 65.8 & 62.3 & \textbf{42.3} & \textbf{4.83} & 4.84 \\
\midrule
\multirow{13}{*}{\rotatebox[origin=c]{90}{\underline{\textbf{Qwen1.5-7B-Chat}}}} & Reference & 65.5 & 23.0 & 34.3 & 69.2 & 59.1 & 74.6 & 75.0 & 57.3 & 6.97 & 10.71 \\
 & AdamW & 69.3 & 22.1 & 38.0 & 71.9 & 59.4 & 76.3 & 71.7 & 58.4 & 7.18 & 11.66 \\

 & \multicolumn{11}{c}{\rule{0pt}{10pt}\multirow{1}{*}[1pt]{\textit{Regularization}}}\\

 & KL & 67.4 & 23.0 & 35.8 & 69.9 & 59.5 & \underline{79.5} & 74.6 & 58.5 & 6.95 & 11.74 \\
 & EMA & 67.3 & 22.6 & 36.1 & 68.9 & 59.0 & 77.8 & 75.4 & 58.2 & 6.86 & 12.92 \\
 & ChildTuning & 66.5 & 23.1 & 35.9 & 70.8 & 58.8 & 78.9 & 74.2 & 58.3 & 7.07 & 11.55 \\
 & LoRA & 67.4 & 22.5 & 35.6 & 69.8 & 59.0 & 78.7 & \textbf{77.0} & 58.6 & 7.04 & 11.33 \\

 & \multicolumn{11}{c}{\rule{0pt}{10pt}\multirow{1}{*}[1pt]{\textit{Offline Merging}}}\\

 & linear & 68.1 & \textbf{24.0} & 35.7 & 69.8 & 59.2 & 77.8 & 72.5 & 58.2 & 6.9 & 10.01\\
 & DARE & 67.5 & 23.3 & 35.9 & 70.2 & 58.9 & 77.2 & \textbf{77.0} & 58.6 & 7.05 & 10.07 \\
 & TIES & 67.1 & \underline{23.8} & 37.8 & 70.5 & 59.3 & \textbf{80.1} & 74.2 & \underline{59.0} & 6.97 & 10.31\\

 & \multicolumn{11}{c}{\rule{0pt}{10pt}\multirow{1}{*}[1pt]{\textit{Online Merging~(Ours)}}}\\
 
 & OnTIES & 67.1 & 22.2 & 38.5 & 72.2 & \textbf{60.4} & 78.7 & 73.0 & 58.9 & 7.1 & 12.57\\
 & OnDARE & 69.4 & 22.9 & \textbf{40.3} & \underline{72.6} & 59.7 & 75.4 & 75.8 & \textbf{59.5} & 7.3 & 11.94 \\
\midrule
\multirow{12}{*}{\rotatebox[origin=c]{90}{\textbf{LLaMa-3-8B-Instruct}}} & Reference & 67.8 & \underline{39.5} & 68.3 & 81.2 & 65.7 & 39.5 & 41.8 & 57.7 & 7.97 & 22.92\\
 & AdamW & 68.3 & 39.3 & 68.4 & \underline{81.7} & 65.7 & 40.9 & 41.4 & 58.0 & 8.04 & 24.10\\

 & \multicolumn{11}{c}{\rule{0pt}{10pt}\multirow{1}{*}[1pt]{\textit{Regularization}}}\\

 & KL & \textbf{70.2} & 38.5 & 67.8 & 81.3 & \textbf{65.8} & 42.1 & 47.1 & \underline{59.0} & 8.12 & 23.08\\
 & EMA & 65.4 & 39.2 & \textbf{69.4} & 81.6 & 65.6 & 42.7 & 43.9 & 58.3 & \underline{8.22} & 24.44\\
 & ChildTuning & 66.8 & 38.0 & 68.6 & 81.0 & 65.0 & 38.9 & \textbf{49.2} & 58.2 & 8.18 & 25.05\\
 & LoRA & \underline{68.4} & 38.7 & 68.2 & 81.3 & \textbf{65.8} & \textbf{45.0} & 43.0 & 58.6 & 8.11 & 23.94\\

 & \multicolumn{11}{c}{\rule{0pt}{10pt}\multirow{1}{*}[1pt]{\textit{Offline Merging}}}\\

 & linear & 67.2 & 39.1 & 68.7 & 80.5 & 65.2 & 40.1 & 36.9 & 56.8 & 8.15 & 24.12 \\
 & DARE & 67.3 & 39.0 & 68.3 & 81.5 & 65.3 & 42.4 & 45.9 & 58.5 & 8.18 & 24.28 \\
 & TIES & 67.1 & \textbf{39.6} & 68.1 & 80.2 & 65.4 & 42.7 & 42.6 & 58.0 & 8.05 & 24.31 \\

 & \multicolumn{11}{c}{\rule{0pt}{10pt}\multirow{1}{*}[1pt]{\textit{Online Merging~(Ours)}}}\\

 & OnDARE & \underline{68.4} & 38.0 & \underline{69.2} & \textbf{82.2} & 65.5 & \underline{43.9} & \underline{48.0} & \textbf{59.3} & \textbf{8.23} & 24.94 \\
 & OnTIES & 68.2 & 38.6 & 68.4 & 80.9 & 65.6 & 43.6 & 41.0 & 58.0 & 8.2 & 25.67\\
    \bottomrule
    \end{tabular}
    }}
    \label{tab:main_results}
    \vspace{-1em}
\end{table}

\begin{sidewaystable}
    \centering
    \small
    \caption{Detailed results on each benchmark of experiments trained from LLaMa-3-8B-It}
    \resizebox{.95\textwidth}{!}{
\begin{tabular}{lrrrrrrrrrrrrrr}
\toprule
                         method &  gsm8k/acc &  math401/acc &  math23k/acc &  ds1000/pass-1 &  mbpp/pass-1 &  codeapex/pass-1 &  humaneval/pass-1 &  ifeval/strict\_prompt\_acc &  ifeval/strict\_instruction\_acc &  coqa/acc &  drop/acc &  mmlu/acc &  agent\_nous/function\_call/acc &  code\_switch\_detect/acc \\
\midrule
Reference &  79.15 &             60.40 &              64.0 &          29.91 &         47.8 &            40.42 &                 40.05 &                     63.49 &                          73.20 &     83.26 &     79.05 & 65.70 &       39.47 &                   41.80 \\
                           AdamW &  79.15 &             61.39 &              64.4 &          26.50 &         49.6 &            40.63 &                 40.62 &                     63.49 &                          73.38 &     84.05 &     79.25 & 65.72 &       40.94 &                   41.39 \\
                             KL &  80.89 &             62.38 &              67.2 &          25.64 &         50.8 &            37.26 &                 40.16 &                     62.85 &                          72.84 &     83.93 &     78.64 & 65.76 &       42.11 &                   47.13 \\
                            EMA &  78.85 &             55.45 &              62.0 &          25.93 &         50.2 &            40.63 &                 40.21 &                     64.23 &                          74.58 &     83.83 &     79.36 & 65.55 &       42.69 &                   43.85 \\
                    ChildTuning &  77.86 &             57.43 &              65.2 &          26.78 &         49.0 &            35.37 &                 40.83 &                     64.23 &                          72.96 &     85.23 &     76.83 & 65.01 &       38.89 &                   49.18 \\
                           LoRA &  79.08 &             60.40 &              65.6 &          26.50 &         49.2 &            38.32 &                 40.62 &                     63.12 &                          73.20 &     83.77 &     78.89 & 65.78 &       45.03 &                   43.03 \\
                         linear &  79.45 &             58.42 &              63.6 &          26.21 &         49.6 &            40.42 &                 40.36 &                     63.86 &                          73.44 &     82.03 &     79.02 & 65.20 &       40.06 &                   36.89 \\
                           DARE &  78.24 &             56.44 &              67.2 &          25.36 &         50.8 &            38.95 &                 40.83 &                     63.59 &                          73.02 &     83.75 &     79.27 & 65.33 &       42.40 &                   45.90 \\
                           TIES &  78.01 &             56.44 &              66.8 &          28.49 &         51.4 &            39.79 &                 38.70 &                     63.03 &                          73.26 &     82.26 &     78.11 & 65.35 &       42.69 &                   42.62 \\
                         OnDARE &  78.85 &             60.40 &              66.0 &          25.07 &         49.4 &            37.89 &                 39.69 &                     64.33 &                          73.98 &     84.79 &     79.52 & 65.47 &       43.86 &                   47.95 \\
                         OnTIES &  79.61 &             61.39 &              63.6 &          27.07 &         47.8 &            39.37 &                 40.36 &                     63.22 &                          73.50 &     83.06 &     78.77 & 65.57 &       43.57 &                   40.98 \\
\bottomrule
\end{tabular}}
    \label{tab:detail_results_llama3}
\end{sidewaystable}

\begin{sidewaystable}
    \centering
    \small
    \caption{Detailed results on each benchmark of experiments trained from Qwen1.5-1.8B-Chat}
    \resizebox{.95\textwidth}{!}{
\begin{tabular}{lrrrrrrrrrrrrrr}
\toprule
                         method &  gsm8k/acc &  math401/acc &  math23k/acc &  ds1000/pass-1 &  mbpp/pass-1 &  codeapex/pass-1 &  humaneval/pass-1 &  ifeval/strict\_prompt\_acc &  ifeval/strict\_instruction\_acc &  coqa/acc &  drop/acc &  mmlu/acc &  agent\_nous/function\_call/acc &  code\_switch\_detect/acc \\
\midrule
                                ChildTuning &  24.56 &             44.55 &              41.6 &           3.42 &         13.6 &            11.79 &                  5.16 &                     17.38 &                          28.48 &     64.37 &     41.01 & 43.89 &       67.54 &                   62.30 \\
                                         KL &  24.11 &             47.52 &              40.8 &           3.70 &         12.2 &            13.47 &                  4.17 &                     17.84 &                          28.36 &     62.89 &     39.68 & 43.48 &       68.71 &                   62.70 \\
                                  Reference &  25.55 &             41.58 &              44.8 &           3.13 &         11.0 &            11.79 &                  5.36 &                     18.11 &                          28.00 &     63.28 &     41.15 & 43.11 &       70.76 &                   58.61 \\
 AdamW &  25.32 &             45.54 &              42.4 &           3.13 &         13.4 &            11.58 &                  5.31 &                     17.93 &                          28.18 &     62.25 &     39.26 & 43.18 &       67.54 &                   61.89 \\
                                       DARE &  23.73 &             40.59 &              42.0 &           4.27 &         12.4 &            10.95 &                  4.84 &                     17.84 &                          28.72 &     62.26 &     36.97 & 43.13 &       73.10 &                   61.07 \\
                                       TIES &  22.74 &             42.57 &              43.2 &           2.85 &         12.8 &            11.58 &                  5.42 &                     18.67 &                          29.14 &     62.61 &     39.37 & 42.64 &       69.01 &                   61.07 \\
                                        EMA &  22.59 &             41.58 &              43.2 &           3.42 &         11.2 &            11.79 &                  5.36 &                     17.38 &                          27.22 &     59.75 &     39.10 & 43.26 &       70.47 &                   63.52 \\
                                     linear &  23.73 &             44.55 &              40.0 &           2.28 &         13.2 &            13.68 &                  4.58 &                     18.48 &                          29.26 &     62.17 &     38.88 & 43.01 &       70.47 &                   61.89 \\
                                       LoRA &  23.65 &             44.55 &              41.2 &           2.28 &         14.2 &            12.84 &                  4.95 &                     18.85 &                          29.02 &     63.38 &     41.32 & 43.51 &       67.84 &                   59.43 \\
                                     OnDARE &  25.25 &             41.58 &              42.8 &           3.13 &         11.6 &            11.16 &                  5.31 &                     19.04 &                          29.44 &     67.89 &     43.54 & 43.96 &       65.79 &                   62.30 \\
                                     OnTIES &  24.34 &             44.55 &              38.4 &           3.99 &         11.4 &            11.37 &                  4.64 &                     19.04 &                          29.14 &     61.50 &     39.11 & 43.44 &       69.01 &                   63.93 \\
\bottomrule
\end{tabular}}
    \label{tab:detail_results_qwen1.8}
\end{sidewaystable}

\begin{sidewaystable}
    \centering
    \small
    \caption{Detailed results on each benchmark of experiments trained from Qwen1.5-7B-Chat}
    \resizebox{.95\textwidth}{!}{
\begin{tabular}{lrrrrrrrrrrrrrr}
\toprule
                         method &  gsm8k/acc &  math401/acc &  math23k/acc &  ds1000/pass-1 &  mbpp/pass-1 &  codeapex/pass-1 &  humaneval/pass-1 &  ifeval/strict\_prompt\_acc &  ifeval/strict\_instruction\_acc &  coqa/acc &  drop/acc &  mmlu/acc &  agent\_nous/function\_call/acc &  code\_switch\_detect/acc \\
\midrule
                             KL &  56.18 &             69.31 &              76.8 &          16.81 &         28.8 &            24.63 &                 21.61 &                     30.31 &                          41.25 &     79.38 &     60.41 & 59.50 &       79.53 &                   74.59 \\
 \textbackslash rowcolor[gray]\{0.95\}Reference &  56.41 &             69.31 &              70.8 &          16.24 &         32.2 &            24.42 &                 19.32 &                     28.93 &                          39.75 &     78.72 &     59.73 & 59.09 &       74.56 &                   75.00 \\
                           AdamW &  58.83 &             67.33 &              81.6 &          16.24 &         28.6 &            22.32 &                 21.30 &                     32.81 &                          43.11 &     81.68 &     62.16 & 59.38 &       76.32 &                   71.72 \\
                           DARE &  58.68 &             68.32 &              75.6 &          15.95 &         30.6 &            26.95 &                 19.64 &                     30.13 &                          41.73 &     80.65 &     59.83 & 58.94 &       77.19 &                   77.05 \\
                           TIES &  57.62 &             65.35 &              78.4 &          16.24 &         33.0 &            25.26 &                 20.62 &                     32.35 &                          43.35 &     79.15 &     61.95 & 59.25 &       80.12 &                   74.18 \\
                            EMA &  57.24 &             67.33 &              77.2 &          14.25 &         31.2 &            25.68 &                 19.22 &                     31.05 &                          41.19 &     80.39 &     57.48 & 58.99 &       77.78 &                   75.41 \\
                         linear &  56.33 &             71.29 &              76.8 &          15.38 &         32.2 &            27.37 &                 21.09 &                     30.13 &                          41.31 &     80.11 &     59.57 & 59.24 &       77.78 &                   72.54 \\
                           LoRA &  58.53 &             66.34 &              77.2 &          16.24 &         29.8 &            25.47 &                 18.39 &                     30.50 &                          40.65 &     79.73 &     59.80 & 59.01 &       78.65 &                   77.05 \\
                         OnDARE &  59.82 &             69.31 &              79.2 &          15.67 &         30.6 &            25.26 &                 20.26 &                     34.75 &                          45.92 &     82.30 &     62.96 & 59.69 &       75.44 &                   75.82 \\
                         OnTIES &  58.38 &             69.31 &              73.6 &          14.53 &         28.6 &            25.05 &                 20.68 &                     32.81 &                          44.12 &     81.89 &     62.50 & 60.41 &       78.65 &                   72.95 \\
\bottomrule
\end{tabular}
}
    \label{tab:detail_results_qwen7}
\end{sidewaystable}

\end{document}